\documentclass[10pt,twocolumn,letterpaper]{article}

\usepackage{cvpr}              
\usepackage[utf8]{inputenc} 
\usepackage[T1]{fontenc}    
\usepackage{url}            
\usepackage{booktabs}       
\usepackage{amsfonts}       
\usepackage{nicefrac}       
\usepackage{microtype}      

\usepackage{graphicx}
\usepackage{multirow}
\usepackage{comment}

\usepackage{enumitem}
\usepackage{tabularx}
\usepackage{adjustbox}
\usepackage{threeparttable}
\usepackage{amsmath}
\usepackage{algorithm}
\usepackage{algorithmic}
\usepackage{multirow}
\usepackage{tabu}
\usepackage{amssymb}
\usepackage{varwidth}
\usepackage{wrapfig}
\usepackage{booktabs}
\usepackage{pifont}
\usepackage[usenames,dvipsnames]{color}

\usepackage{subcaption}
\usepackage{transparent}
\usepackage{tabulary}
\usepackage{amsfonts}
\usepackage{lipsum}

\usepackage{times}
\usepackage{epsfig}
\usepackage{graphicx}
\usepackage{amsmath}
\usepackage{amssymb}

\newcommand{\figref}[1]{Fig.~\ref{#1}}
\newcommand{\tabref}[1]{Table.~\ref{#1}}

\newcommand{\cmark}{\textcolor{green}{\ding{51}}}%
\newcommand{\xmark}{\textcolor{red}{\ding{55}}}%

\usepackage{times}
\usepackage{epsfig}
\usepackage{graphicx}
\usepackage{amsmath}
\usepackage{amssymb}
\usepackage{textpos}

\newcommand\blfootnote[1]{%
  \begingroup
  \renewcommand\thefootnote{}\footnote{#1}%
  \addtocounter{footnote}{-1}%
  \endgroup
}

\usepackage[dvipsnames]{xcolor}

\definecolor{cvprblue}{rgb}{0.21,0.49,0.74}
\usepackage[pagebackref,breaklinks,colorlinks,citecolor=cvprblue]{hyperref}

\def\paperID{8826} 
\def\confName{CVPR}
\def\confYear{2024}

\begin{document}

\twocolumn[{%
\renewcommand\twocolumn[1][]{#1}%
\title{
       MTMMC: A Large-Scale Real-World Multi-Modal Camera Tracking Benchmark
       }  
\author{
Sanghyun Woo$^{1*}$ \quad
Kwanyong Park$^{2*}$ \quad
Inkyu Shin$^{3*}$ \quad
Myungchul Kim$^{3*}$ \quad
In So Kweon$^3$ \\
\\
$^1$New York University \quad \quad \quad
$^2$ETRI \quad \quad \quad
$^3$KAIST
}
\maketitle

\begin{center}
    \centering
    \vspace{-3mm}
    \includegraphics[width=0.98\textwidth]{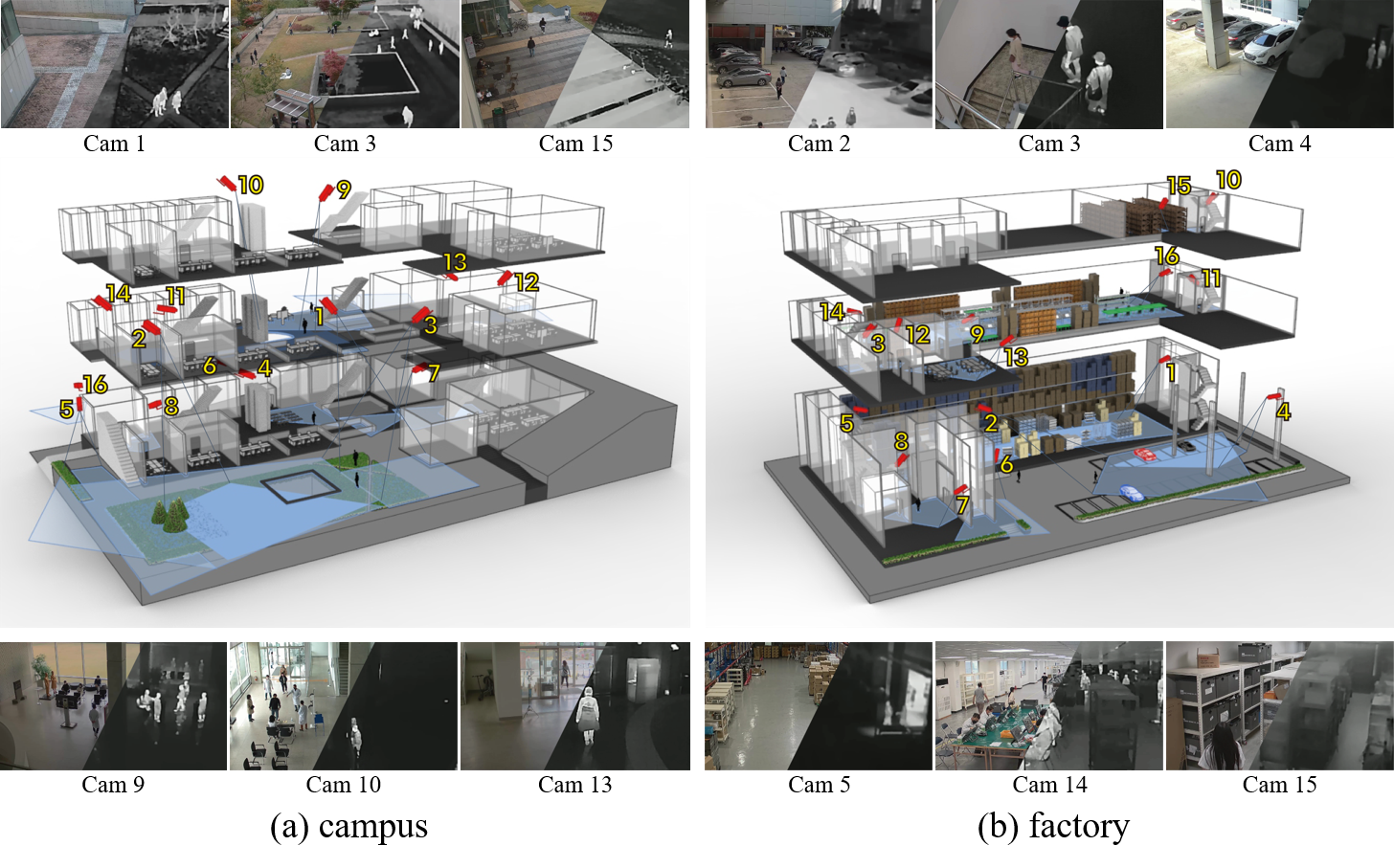}
    \vspace{-3mm}
    \captionsetup{font=footnotesize}
    \captionof{figure}{
    \textbf{The 3D layout overview.} 
    (a) campus and (b) factory.
    We installed 16 multi-modal cameras in both indoor and outdoor settings, across multiple floors, with overlapping coverage. The cameras were fixed in position and angle to densely cover the building, creating a realistic surveillance camera system. 
    }
    \label{fig:main_figure}
\end{center}%
}]

\blfootnote{* Equal contribution}

\begin{textblock*}{.8\textwidth}[.5,0](0.5\textwidth, -.72\textwidth)
\centering
{\hspace{-0ex} \small \url{https://sites.google.com/view/mtmmc}}
\end{textblock*}


\maketitle
\begin{abstract}
\vspace{-1em}
Multi-target multi-camera tracking is a crucial task that involves identifying and tracking individuals over time using video streams from multiple cameras. This task has practical applications in various fields, such as visual surveillance, crowd behavior analysis, and anomaly detection. However, due to the difficulty and cost of collecting and labeling data, existing datasets for this task are either synthetically generated or artificially constructed within a controlled camera network setting, which limits their ability to model real-world dynamics and generalize to diverse camera configurations.
To address this issue, we present MTMMC, a real-world, large-scale dataset that includes long video sequences captured by 16 multi-modal cameras in two different environments - campus and factory - across various time, weather, and season conditions. This dataset provides a challenging test-bed for studying multi-camera tracking under diverse real-world complexities and includes an additional input modality of spatially aligned and temporally synchronized RGB and thermal cameras, which enhances the accuracy of multi-camera tracking. MTMMC is a super-set of existing datasets, benefiting independent fields such as person detection, re-identification, and multiple object tracking. We provide baselines and new learning setups on this dataset and set the reference scores for future studies. 
The datasets, models, and test server will be made publicly available.
\vspace{-1em}
\end{abstract}    
\vspace{-1em}
\section{Introduction}
\label{sec:intro}

Multiple object tracking (MOT) is an essential vision task that helps us understand visual content and predict the evolution of the surroundings over time.
Recent advancements in MOT, thanks to benchmarks such as MOT17~\cite{milan2016mot16}, BDD100K~\cite{yu2020bdd100k}, Waymo~\cite{sun2020scalability}, and TAO~\cite{dave2020tao} have led to the development of more effective and efficient trackers~\cite{feichtenhofer2017detect,zhou2020tracking,meinhardt2021trackformer,pang2021quasi,woo2022tracking}. 
Despite these advancements, multiple-camera tracking has seen limited exploration, largely due to the lack of appropriate datasets. 
The high costs associated with the collection and annotation of such data are a major bottleneck.

The datasets currently available predominantly consist of either synthetically generated data from game simulators~\cite{kohl2020mta} or small-scale real-world data obtained from controlled camera networks~\cite{fleuret2007multicamera,de2008distributed,d2009semi,ferryman2009pets2009,berclaz2011multiple,cao2015equalised}, which assume an idealized overlap between the camera views to simplify the annotation process.
However, synthetic data often fail to translate effectively to real-world scenarios due to significant domain shifts, and datasets from controlled environments do not reflect the complexities of real-world multi-camera networks.
Additionally, the withdrawal of the DukeMTMC~\cite{ristani2016performance}, previously the most extensive real-world dataset, due to privacy issues has left a considerable void in this research area.

To tackle this, this paper presents a new benchmark called the Multi-Target Multi-Modal Camera (MTMMC) tracking dataset. The dataset was collected from two challenging environments—a campus and a factory—equipped with 16 multi-modal cameras, each placed at different angles (see~\figref{fig:main_figure}). The dataset consists of 25 video recordings—13 from the campus and 12 from the factory—with each video containing five and a half minutes of HD video recording captured under various times, weather, and seasons, ensuring a rich diversity of backgrounds. 
To ensure compliance with data privacy standards, we collected informed consent from all participants, who explicitly agreed to the public release of the collected data for research purposes. 
The annotation of all trajectories was accomplished using a semi-automatic labeling system, carefully refined by crowdworkers over \emph{a year}, making the dataset the most largest publicly accessible MTMC tracking benchmark to date.

Significantly, our dataset contains both RGB and thermal cameras, allowing the tracker to additionally utilize thermal information for more accurate multi-camera tracking. This is the first time a dataset has provided a valid test-bed for studying the impact of multi-modal learning for multi-camera tracking. Our experiments reveal that incorporating thermal data into standard RGB camera-based trackers results in more robust tracking, motivating future research in this new direction. 
The construction of the MTMC dataset also facilitates progress in related subtasks, such as person detection, re-identification, and MOT. 


\begin{table*}[t!]
\centering
\resizebox{\linewidth}{!}{
\begin{tabular}{l c c c c c c c c}
\toprule
Dataset                                  & \# Cameras & \# ID  & \# Frames & OV/NOV     & Camera Coverage  & Extra Modality  & FPS & Resolution \\ \midrule
PETS2009~\cite{ferryman2009pets2009}     & 8          & 30     & 1,200     & OV         & outdoor          & \xmark         & 30   & $768\times 576$ \\
USC Campus~\cite{kuo2010inter}           & 3          & 146    & 135,000   & NOV        & outdoor          & \xmark         & 30  & $852\times 480$ \\                                 
Passageway~\cite{berclaz2011multiple}    & 4          & 4      & 120,000   & OV         & outdoor          & \xmark         & 25  & $320\times 240$ \\
NLPR MCT~\cite{cao2015equalised}         & $\leq5$    &$\leq235$ & 355,500   & NOV        & in \& outdoor    & \xmark         & 20  & $320\times 240$ \\
CamNet~\cite{zhang2015camera}            & 8          & 50     & 360,000   & NOV        & in \& outdoor    & \xmark         & 25  & $640\times 480$ \\
WILDTRACK~\cite{chavdarova2017wildtrack} & 7          & N/A    & 66,626    & both       & outdoor          & \xmark         & 60  & $1920\times 1080$ \\
DukeMTMC~\cite{ristani2016performance}   & 8          & 2,834  & 2,448,000 & NOV        & outdoor          & \xmark         & 60  & $1920\times 1080$ \\
MTA~\cite{kohl2020mta}                   & 6          & 2,840  & 2,007,360 & both       & simulated        & \xmark         & 41  & $1920\times 1080$ \\
MMPTRACK~\cite{han2021mmptrack}          & $\leq6$    & $\leq140$ & 2,979,900 & OV    & indoor           & \xmark         & 15  & $640\times 320$ \\
\midrule
MTMMC (Ours)     &  \textbf{16}        & \textbf{3,669}  & \textbf{3,052,800} & \textbf{both}  & \textbf{in} \& \textbf{outdoor}    & \cmark~(\textbf{Thermal})& 23  & \textbf{1920} $\times$ \textbf{1080} \\
\bottomrule
\end{tabular}
}
\captionsetup{font=footnotesize}
\caption{
\textbf{Overview of the publicly available MTMC datasets.}
For each dataset, we report the number of cameras, person identities, and frames.
We also report the presence of overlapping (OV) / non-overlapping (NOV) camera views, camera coverage,
availability of extra input modality, annotated frame rate (FPS), and frame resolution.
Our new MTMMC dataset is unprecedented in its scale and diversity. It includes 16 cameras, 3,669 person IDs, and 3 million frames, making it a challenging and large-scale dataset. The dataset also provides high-resolution and multi-modal information.
}
\label{tab:MTMC_stats}
\end{table*}
\section{Related Work}

\noindent\textbf{Benchmarks}
To construct a high-quality MTMC dataset, it is crucial to have temporally synchronized videos from multiple cameras. These videos must also maintain consistent person identities across all camera views.
However, this requirement results in high annotation costs. As a result, existing MTMC benchmarks are either short in duration~\cite{fleuret2007multicamera,de2008distributed,d2009semi,ferryman2009pets2009}, have low video resolution~\cite{berclaz2011multiple,kuo2010inter,zhang2015camera,cao2015equalised}, or provide inconsistent person IDs~\cite{chavdarova2017wildtrack}, making them unsuitable for training generic deep trackers for real-world use cases.
The most popular dataset closest to our proposal is DukeMTMC~\cite{ristani2016performance}, but it was withdrawn due to consent and privacy issues.
Recently, two large-scale MTMC datasets, MTA~\cite{kohl2020mta} and MMPTRACK~\cite{han2021mmptrack}, have been introduced, but they have limitations such as being obtained through game simulations or collected in controlled setups where all cameras have overlapping fields of view. 
The new MTMC dataset aims to provide a larger basis for training and testing MTMC performance than any previous datasets, making it a valuable resource for researchers.
\\
\\
\noindent\textbf{Multi-modal Learning}
Unlike existing tracking datasets, our dataset features an additional thermal input modality, which opens up new research directions for multi-modal learning in multi-camera tracking. Multi-modal learning is an interesting research problem that not only improves model robustness through modality fusion, which applies to various vision tasks such as detection~\cite{li2019illumination,zhang2019cross,zhang2019weakly,zhang2019weakly,zhang2019weakly}, visual object tracking~\cite{yan2021depthtrack,kart2018make,kart2019object,liu2018context,kristan2020eighth}, and segmentation~\cite{sun2019rtfnet,zhou2021gmnet}, but also enables better representations of each modality by learning the intrinsic correlations between them~\cite{xu2017learning,luo2018graph,zhao2020knowledge,abavisani2019improving,shin2023self}. In this paper, we present two new experimental setups with baselines.
\\
\\
\noindent\textbf{Multiple Object Tracking}
The standard way to tackle the MTMC problem involves a two-step approach:
1) generating local tracklets for all the targets within each camera;
2) associating these local tracklets across cameras when they belong to the same target.
The first step, known as multiple object tracking, has been extensively studied by the community. The tracking-by-detection paradigm has emerged as the dominant approach, owing to significant improvements in object detection techniques~\cite{ren2015faster,lin2017focal,redmon2017yolo9000,lin2017feature}.
Recent advances in this paradigm include developing more discriminative association objectives~\cite{hu2019joint,leal2016learning,park2022per,wang2020towards,wojke2017simple,pang2021quasi,woo2022bridging}, 
unifying detection and tracking~\cite{feichtenhofer2017detect,sun2019deep,wu2021track,zhou2020tracking}, or building an end-to-end framework~\cite{sun2020transtrack,meinhardt2021trackformer,zeng2021motr}.
\\
\\
\noindent\textbf{Multi Camera Association}
Cross-camera association presents a more challenge due to pronounced changes in object appearance between cameras, variable background conditions, and an increased number of targets to be matched. 
To facilitate this process, various constraints have been employed, including time conflicts~\cite{zhang2017multi}, linear motion patterns~\cite{ristani2018features}, camera network topology~\cite{shiva2017distributed,jiang2018online}, geometric cues~\cite{bredereck2012data,you2020real,chen2020cross}, and spatial locality~\cite{hou2021adaptive}. 

Cross-camera association can be conducted in both real-time~\cite{cao2015equalised,quach2021dyglip} and offline manners~\cite{das2014consistent,zhang2017multi,ristani2018features,he2020multi,hou2021adaptive}, with the latter often favored for its enhanced accuracy.
Notably, several offline global association techniques have been developed, such as hierarchical clustering~\cite{zhang2017multi,zhong2017re}, correlation clustering~\cite{ristani2018features,bonchi2014correlation}, matrix factorization~\cite{he2020multi}, and adaptive locality-based association~\cite{hou2021adaptive}.


\begin{figure*}[!t]
  \includegraphics[width=\linewidth]{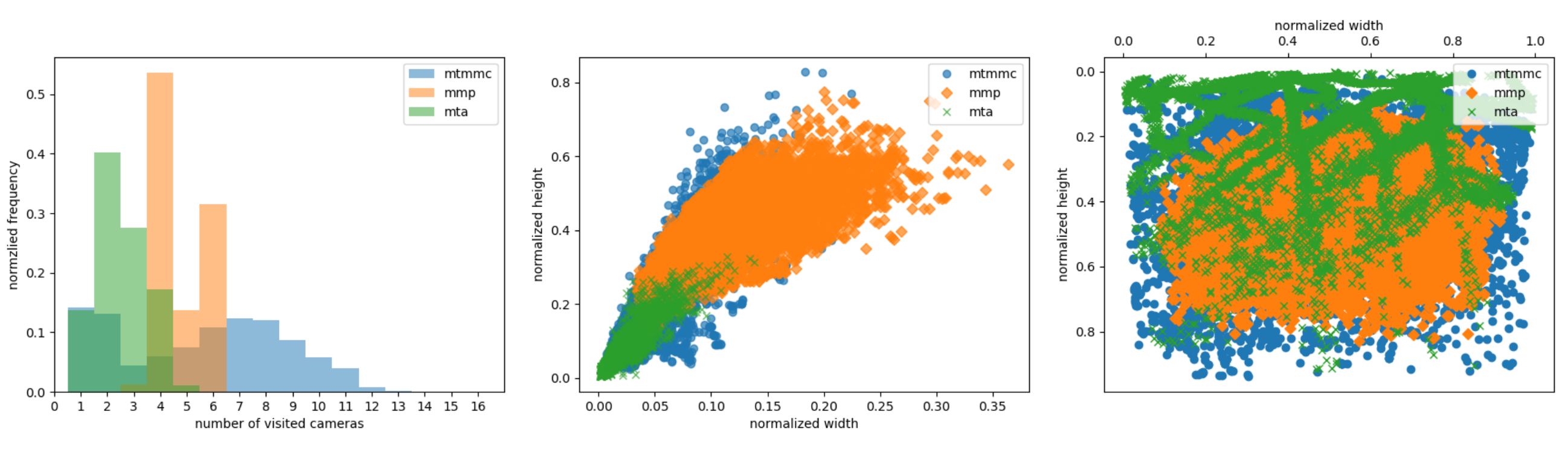}
  \vspace{-6mm}
  \captionsetup{font=footnotesize}
  \caption{\textbf{MTMC Dataset statistics comparison.} We compare \textcolor{RoyalBlue}{\textbf{MTMMC}} dataset with the current largest \textcolor{OliveGreen}{\textbf{simulated MTA}}~\cite{kohl2020mta} and \textcolor{Orange}{\textbf{real world MMPTrack}}~\cite{han2021mmptrack} datasets.
  Each visual summarizes specific statistics of each dataset :
  (a) The instance tracks by the number of visited cameras,
  (b) The joint distribution of instance normalized width and height,
  (c) The instance track centers plotted over normalized image coordinates.
  }
  \vspace{-3mm}
  \label{fig:overview}
\end{figure*}

\section{MTMMC} 
Our camera network is detailed in~\figref{fig:main_figure}, 
comprising a school campus and a factory environment.
This choice reflects common real-world surveillance scenarios.
With 16 multi-modal cameras, our configuration spans both indoors and outdoors, extends over multiple levels of floors. 
Each camera provides RGB and thermal data that are spatially aligned and temporally synchronized.
A detailed description of all cameras is provided in the appendix.

\subsection{Data Characteristics}
\tabref{tab:MTMC_stats} presents a comparative summary between our dataset and the existing datasets.
The MTMMC dataset consists of 25 scenarios, each composed of 16 \textbf{high-resolution} RGB~+~Thermal videos captured at 23 fps, both at \textbf{indoor and outdoor}, resulting in a total of 3,052,800 frames.
The dataset offers diverse real-world environmental conditions, ranging from day to evening (\emph{time}), sunny to cloudy (\emph{weather}), and summer to fall (\emph{season}). 
This \textbf{diversity} makes our dataset unique and more representative.

In \figref{fig:overview}, we present a detailed comparison of our MTMMC dataset with two of the largest datasets in MTMC: MTA~\cite{kohl2020mta} and MMPTrack~\cite{han2021mmptrack}.
We focus on three critical aspects that influence the tracking performance:

\begin{enumerate}
\itemsep0.5em
    \item \textbf{Track Length.} We analyze the variability in the number of cameras each individual is tracked through, providing insights into the robustness of tracking across cameras. MTMMC features a broader distribution of track lengths, including more extended tracking periods compared to MTA and MMPTrack, as shown in Figure~\ref{fig:overview}(a), demonstrating our dataset's capability to \emph{challenge and train models on maintaining identities over longer sequences}.
    
    \item \textbf{Track Scale.} We assess the range of target scales by normalizing bounding boxes against image dimensions. MTMMC includes a diverse range of scales, with Figure~\ref{fig:overview}(b) highlighting instances of small-scale tracks that are not well-represented in other datasets, critical for \emph{training models to detect and track small or distant targets}.
    
    \item \textbf{Track Path.} Our analysis of track trajectories in normalized image coordinates reveals MTMMC's coverage of diverse movement patterns. The dataset exhibits a more comprehensive range of trajectories compared to others, as depicted in Figure~\ref{fig:overview}(c), which is pivotal for algorithms \emph{predicting and maintaining track continuity amidst complex environmental dynamics}.
\end{enumerate}

\noindent MTMMC advances in all these three aspects over the previous datasets, providing a challenging testbed that more precisely reflects real-world conditions.

We further provide a statistical analysis of key attributes of our dataset in the appendix.
First, 
the number of objects per frame and the number of tracks per video metrics reveal the \textbf{complexity} of scene contexts and the robustness required for successful multi-object tracking.
Additionally, the \textbf{wide demographic range}, represented by the age and gender distributions of the actors ensures the development of more inclusive and unbiased tracking algorithms.
Lastly, for the first time, a thermal modality is included, enabling \textbf{multi-modal} learning — a feature unprecedented in previous multi-object tracking datasets.

%



\subsection{Data Collection}
The data collection was conducted over two days, capturing different seasons for each environment: summer for the factory and fall for the school campus. 
To ensure a high degree of accuracy in temporal alignment, we employed a precise global time-stamping method for space-time synchronization. For potential frame drops, we meticulously inspected the
video sequences and made adjustments by aligning timestamps and interpolating missing data.
To prevent any privacy issues, we recruited 623 actors of varying ages and genders and obtained data release agreements. We ensured that all participating actors were compensated for their time and efforts. Furthermore, we conducted de-identification for the 107 non-actors involved in the recordings.

Each video last in five-and-a-half-minutes per scenario, with 12 scenarios from the factory and 13 scenarios from the school campus. We allowed the actors to improvise their actions, provided they fit the given circumstances. For instance, actors could move luggage in the factory or play ball at school, resulting in a wide variety of behaviors being captured.
Moreover, we instructed the actors to change their clothes for each scenario to ensure diverse appearances.

Notably, our new MTMMC dataset significantly improves upon the Duke-MTMC dataset~\cite{ristani2016performance}, which was collected within a narrow 1.5-hour window on a single day on campus. By extending the breadth and diversity of our collection process, we aim to provide a more solid foundation for the development of robust tracking systems.

\subsection{Data Annotation}

We designed an annotation pipeline to separate the single-camera tracking and the multi-camera association tasks. The single-camera tracking involves generating bounding boxes of person tracks within a camera, while the multi-camera association involves assigning consistent person-IDs across multiple cameras.
By dividing the tasks, we can assign inexperienced annotators to the former and skilled workers to the latter. 
The reviewers carefully check the quality of the completed labels from the annotators, and this process is repeated several times until no critical errors are visible.
More details are in the appendix. 

\vspace{3mm}
\noindent\textbf{Single Camera Tracking.}
To annotate the set of 400 videos (16 cameras $\times$ 25 scenarios), we tasked annotators with drawing bounding boxes and assigning track-IDs to each person in the video. We collected the annotations in a semi-automatic manner, as described in previous works~\cite{weber2021step,wang2021unidentified}.
First, the annotators tracked and labeled the person in the keyframes, which were selected every five frames in a video. We then used the deepSORT~\cite{wojke2017simple} algorithm to generate pseudo tracking boxes by interpolating the annotations between keyframes efficiently. The predicted tracking boxes were then carefully corrected by the annotators.
Additionally, to protect the privacy of non-actors, we applied a de-identification process, which involved blurring their faces while remaining the ground truth annotations intact. This process ensured confidentiality of personal information, while simultaneously preserving the data integrity. 

\vspace{3mm}
\noindent\textbf{Multiple Camera Association.}
In the next step, we asked annotators to assign consistent track-IDs for the same person \textit{across the cameras} for each scenario. We observed that the semi-automatic labeling approach was not sufficient to achieve satisfactory label quality for this task. Hence, we relied on careful manual labeling.
After the initial labeling was completed by the annotators, the reviewers collected person-ID errors using two critical camera constraints. Firstly, one person cannot appear in multiple tracks of the same camera simultaneously. Secondly, one person cannot be visible in the view of two non-overlapping cameras simultaneously. 
The reviewers also checked for other remaining errors.
All the collected errors were then passed to the annotators, who corrected them. 
The refining process was iterated twice to guarantee high-quality labels. 

\subsection{Data Splits}
We split the MTMMC dataset into three subsets. The \textbf{train} set includes 14 scenarios (7 from the factory and 7 from the campus), the \textbf{validation} set includes 5 scenarios (3 from the factory and 2 from the campus), and the \textbf{test} set includes 6 scenarios (2 from the factory and 4 from the campus).

\section{Experiments}

We present various experimental setups and benchmark their performance using the new MTMMC dataset. 
For the evaluation, we use standard metrics such as mAP for \emph{detection}, Rank1 and mAP for \emph{re-identification}, and CLEAR MOT and IDF1 for \emph{tracking}. The experiments are conducted on the \textbf{train} and \textbf{validation} sets of the dataset. 
Detailed setup specifications are in the appendix.

\subsection{Sub Tasks: Detection, Re-ID, and MOT}\label{subsec:benchmark_result}

\paragraph{Person Detection}
We evaluate the efficacy of our dataset in training person detectors for tracking applications.
We utilized two well-established detectors, Faster RCNN~\cite{ren2015faster} and YOLOX~\cite{ge2021yolox}, and investigate how well these models generalize and perform when trained on task-specific versus generic datasets.
Specifically, we trained models MTMMC-Person and COCO-Person datasets and then tested their generalization performance using the MOT17~\cite{milan2016mot16} dataset, which presents a variety of real-world tracking scenarios.
The COCO-Person is a subset of the larger COCO~\cite{lin2014microsoft} dataset and includes 65K natural images that depict humans.
To compare fairly, we matched the size of our MTMMC-Person dataset, compiling 60K images sampled at a frame rate of 23 fps from the original video footage.

As shown in \tabref{tab:det}, models trained on the MTMMC-Person dataset consistently outperformed those trained on COCO-Person during the MOT17 evaluations.
This suggests that the specificity of the training data to the end-use scenario is crucial.
By design, the MTMMC dataset is tailored to tracking, highlighting diverse human activities, frequent occlusions, varied interactions and non-central camera angles, which are typical in real-world tracking situations.
These results validate the importance of contextual alignment between training data and its target application, emphasizing the value of our specialized dataset, MTMMC, for tracking and surveillance applications.

\begin{table}[t]
\small
\setlength{\tabcolsep}{5pt}
\centering
{
        \def\arraystretch{1.1}
        \begin{tabular}{c|cc|c}
        \hline
        Method & Train on & Eval on & mAP  \\
        \hline
        \multirow{2}{*}{Faster RCNN} & COCO-Person  & MOT17 & 29.8\\
                                     & MTMMC-Person & MOT17 & 31.3 \\
        \cline{1-4}
        \multirow{2}{*}{YOLOX}       & COCO-Person  & MOT17 & 34.2 \\
                                     & MTMMC-Person & MOT17 & 38.3 \\
        \hline
        \end{tabular}
}
\vspace{-2mm}
\caption{\textbf{Detection Results.}}
\vspace{-3mm}
\label{tab:det}
\end{table}



\begin{table}[t]
\small
\setlength{\tabcolsep}{5pt}
\centering
{
        \def\arraystretch{1.1}
        \begin{tabular}{c|cc|cc}
        \hline
        Method & Train on & Eval on & Rank 1 & mAP  \\
        \hline
        \multirow{5}{*}{AGW} & Market-1501 & Market-1501 & 95.3 & 88.2 \\
                             & MSMT17 & MSMT17 & 78.3 & 55.6 \\
                             & MTMMC-reID & MTMMC-reID & 76.0 & 45.6 \\
        \cline{2-5}
                             &MSMT17 & Market-1501 & 64.3 & 34.2 \\
                             &MTMMC-reID & Market-1501 & 66.5 & 35.4 \\
        \hline
        \end{tabular}
}
\vspace{-2mm}
\caption{\textbf{Re-Identification Results.}}
\vspace{-3mm}
\label{tab:re_id}
\end{table}

\paragraph{Person Re-Identification} 
In line with the standard protocols for re-identification (Re-ID) data construction, as outlined in ~\cite{zheng2015scalable,zheng2017unlabeled}, we derived our MTMMC-reID dataset from the larger MTMMC dataset. For our experiments, we used the AGW model~\cite{ye2021deep} as the benchmark.

Re-ID tasks require the identification of individuals across multiple camera views and at different times.
Training data characteristics significantly influence the performance of Re-ID systems. The MTMMC-reID dataset, in particular, provides a challenging training environment, as evidenced by the lower Rank-1 accuracy and mAP scores—76.0 and 45.6, respectively—compared to other datasets (see~\tabref{tab:re_id}, rows 2-4). These figures highlight the demanding nature of the tracking scenarios within MTMMC-reID.

However, the dataset's complexity is beneficial for model generalization. 
For instance, when a model trained on the MSMT17~\cite{wei2018person} dataset is evaluated on Market-1501~\cite{zheng2015scalable}, performance drops (to 64.3 Rank-1 and 34.2 mAP), indicating a loss of generalizability. Yet, if the same model is trained on MTMMC-reID and tested on Market-1501, it demonstrates better robustness with higher Rank-1 accuracy and mAP scores (66.5 and 35.4, respectively) compared to the MSMT17 training (refer to~\tabref{tab:re_id}, rows 5-6). These results imply that despite the intrinsic challenges of MTMMC-reID, models trained on it are better equipped to handle new, unseen environments, underscoring the value of rigorous training environments for improved real-world applicability.

\begin{table*}[t!]
\centering
\resizebox{0.97\linewidth}{!}{
\def\arraystretch{1.2}
\begin{tabular}{c|ccc|ccccc|ccccc}
\toprule
\multicolumn{1}{c|}{\multirow{2}{*}{Method}} & \multicolumn{3}{c|}{Train on} & \multicolumn{5}{c|}{Eval on MTMMC}  & \multicolumn{5}{c}{Eval on MOT17} \\ \cline{2-4}\cline{5-14} 
\multicolumn{1}{c|}{}                         & MTMMC & MOT17 & Misc & IDF1 & MOTA & FP & FN & IDs   & IDF1 & MOTA & FP & FN & IDs \\ \midrule
\multirow{3}{*}{JDE}    & \checkmark       &             &        &    42.4   &  74.6        & 146678  & 859893     & 30767  & 48.0 & 40.9  & 2311   & 29084 & 329      \\
\multicolumn{1}{c|}{}                        &                  & \checkmark  & cccpe     &    34.0 &  52.3        & 206112  & 1694301    & 27347  & 63.6 & 60.0  & 2927   & 18155 & 486      \\
\multicolumn{1}{c|}{}                        & \checkmark       & \checkmark  & cccpe     &    43.7 &  72.6        & 125770  & 964863     & 25725  & 70.5 & 65.7  & 2232   & 15759 & 469      \\ \midrule
\multirow{3}{*}{QDTrack}                  & \checkmark       &             &          &    53.0 &  84.5        & 157529  & 475242     & 14542  & 55.3 & 43.6  & 10548   & 80197 & 449      \\
                                             &                  & \checkmark  &          &    34.3  &  52.3        & 286382  & 1643818    & 21470  & 66.8 & 65.3  & 9324   & 45441 & 1383     \\
                                             & \checkmark       & \checkmark  &           &    54.2 &  84.6        & 439646  & 439646     & 14106  & 70.0 & 68.6  & 6927   & 42903 & 1005     \\ \midrule
\multirow{4}{*}{CenterTrack}                 & \checkmark       &             &           &    50.8 &  78.6        & 504642  & 353525     & 16972  & 55.0 & 45.3  & 17718  & 69870 & 903      \\
                                             &                  & \checkmark  &           &    25.2 &  37.0        & 629624  & 1911628    & 40656  & 62.1 & 60.5  & 6678   & 55446 & 1710     \\
                                             &                  & \checkmark  & CH$_{pre}$& 27.1 & 45.7                     & 518692         & 1662554            & 40746        & 63.7 &  66.2        & 7128  & 45939  & 1611    \\
                                             & \checkmark       & \checkmark  & CH$_{pre}$& 51.6 & 80.9                      & 415132         & 351162            & 16938       & 65.7  &   66.7        & 6138  & 46338 & 1407 \\ \midrule
\multirow{4}{*}{ByteTrack}                   & \checkmark       &             &           & 64.8 &  89.7        & 112835  & 300354     & 7153   & 69.1 & 55.9  & 16896  & 54106 & 230     \\
                                             &                  & \checkmark  &           &    40.2 & 56.8      & 506286  & 1283368    & 13585    & 76.8  & 75.0  & 4539  & 8693  & 224      \\
                                             &                  & \checkmark  & CH        &    56.9 &  77.7        & 267550  & 640084     & 7547  & 79.5 & 76.6  & 10128  & 27250 & 479     \\
                                             & \checkmark       & \checkmark  & CH        &    64.6 &  89.1        & 147385  & 289854     & 7184   & 78.7 & 76.9  & 8504   & 28302 & 517     \\  \bottomrule

\end{tabular}
}
 \captionsetup{font=footnotesize}
 \caption{\textbf{Multi Object Tracking Results.}
 Following the previous works, we use additional person detection data: CH denotes CrowdHuman~\cite{shao2018crowdhuman},
 MIX indicates combined datasets of Caltech Pedestrian~\cite{pedestrian}, Citypersons~\cite{zhang2017citypersons}, CUHK-SYS~\cite{xiao2017joint}, PRW~\cite{zheng2017person} and ETH~\cite{eth_paper}.}
\label{table:MOT}
\end{table*}

\begin{table*}[t]
\setlength{\tabcolsep}{2.5pt}
 \centering
    \subfloat[\scriptsize w/o finetune]
         {
         \resizebox{0.49\textwidth}{!}
        {
        \def\arraystretch{1.3}
        \begin{tabular}{l|cc|ccccc}
        \toprule
        \multicolumn{1}{c|}{\multirow{2}{*}{Method}} & \multicolumn{2}{c|}{Train on} & \multicolumn{5}{c}{Eval on MOT17} \\ \cline{2-8} 
        \multicolumn{1}{c|}{}                        & MTMMC        & MOTSynth                                 & IDF1  & MOTA  & FP     & FN     & IDs    \\ \midrule
        \multirow{3}{*}{QDTrack}                  & \checkmark            &                                 & 55.3 & 43.6       & 10548    & 80197 & 449   \\
                                                     &              & \checkmark                               & 54.1 & 43.1            & 11178       & 80178       & 615 \\
                                                     & \checkmark            & \checkmark                      &  60.8 & 48.9    &    14724     &  67029       & 870         \\ \bottomrule
        \end{tabular}
        }
         
         }
    \subfloat[\scriptsize w/ finetune]
         {
         \resizebox{0.49\textwidth}{!}
        {
        \def\arraystretch{1.3}
        \begin{tabular}{l|cc|ccccc}
        \toprule
        \multicolumn{1}{c|}{\multirow{2}{*}{Method}} & \multicolumn{2}{c|}{Train on}  & \multicolumn{5}{c}{Eval on MOT17} \\ \cline{2-8} 
        \multicolumn{1}{c|}{}                        & MTMMC        & MOTSynth                                 & IDF1  & MOTA  & FP     & FN     & IDs    \\ \midrule
        \multirow{3}{*}{QDTrack}                  & \checkmark            &                      & 68.6 & 66.6        & 9963      &43074       &957       \\
                                                     &                 & \checkmark                 & 70.8 & 68.7           &    9813   & 39882       &921       \\ 
        \multicolumn{1}{c|}{}                        &  \checkmark            & \checkmark         & 72.0 & 70.2           & 8367         & 39135          & 750         \\\bottomrule
        \end{tabular}
        }
         
         }
 \vspace{-3mm}
 \captionsetup{font=footnotesize}     
 \caption{\textbf{Pre-Training} on MTMMC and MOTSynth.}
\label{tab:mot_synth}
 \vspace{-3mm}
\end{table*}

\paragraph{Multi Object Tracking}

Multi-object tracking (MOT) is a task that requires the detection and tracking of multiple objects, often people, through a sequence of video frames. The challenge lies in keeping consistent object identities despite movement, occlusions, and environmental changes. 
In our experiment, we employed four state-of-the-art trackers: JDE~\cite{wang2019towards}, QDTrack~\cite{pang2021quasi}, CenterTrack~\cite{zhou2020tracking}, and ByteTrack~\cite{zhang2021bytetrack}, and our analysis focuses on three main aspects:

\begin{enumerate}
\itemsep0.5em
\item\textbf{Training and Evaluating on the Same Dataset}:
When models are both trained and evaluated on the same dataset, they exhibit lower performance on the MTMMC compared to the MOT17 dataset. For instance, JDE, achieved an IDF1 score of 42.4\% on MTMMC, whereas the same model yielded an improved IDF1 of 63.6\% on MOT17. 
This trend is consistent across all tested models, indicating that MTMMC presents a more challenging testbed.

\item\textbf{Training and Evaluating on Different Datasets}:
When training and evaluation datasets differed, we observed a pattern where models trained on MTMMC generally outperformed those trained on MOT17 when evaluated on the alternate dataset. 
For example, ByteTrack, after being trained on MTMMC and tested on MOT17, reached an IDF1 score of 69.1\% and MOTA of 55.9\%, which is closer to the practical upper bounds observed when trained and tested on MOT17 (IDF1 of 76.8\% and MOTA of 75.0\%).
In contrast, when ByteTrack was trained on MOT17 and evaluated on MTMMC, it achieved a much lower IDF1 of 40.2\% and MOTA of 56.8\%, versus its upper-bound performance on MTMMC (IDF1 of 64.8\% and MOTA of 89.7\%). This suggests that the complex and diverse tracking environments found in MTMMC contribute to the development of more robust and generalizable model features. 

\vspace{1mm}

Notably, the above two trends within multi-object tracking mirror the tendencies observed in our Re-ID experiments. This consistency reinforces the notion that training on more complex and diverse environments effectively enhances the models' ability to generalize and maintain accuracy when introduced to new domains.

\item\textbf{Training on Combined Datasets}:
The most compelling results were observed when models were trained on a mixture of both MTMMC and MOT17 datasets. This combined training approach produced the best results on both MTMMC and MOT17 evaluations. It implies that the MTMMC provides a complementary training signal
to the MOT17. 
When combined, the diversity and complexity of MTMMC complement the MOT17, leading to a robust tracking model. 
\end{enumerate}

In conclusion, these experiments underline the importance of dataset diversity and complexity in training multi-object tracking models. The demanding context provided by MTMMC help to forge models that can handle real-world complexities effectively.



\begin{figure*}[!t]
  \includegraphics[width=\linewidth]{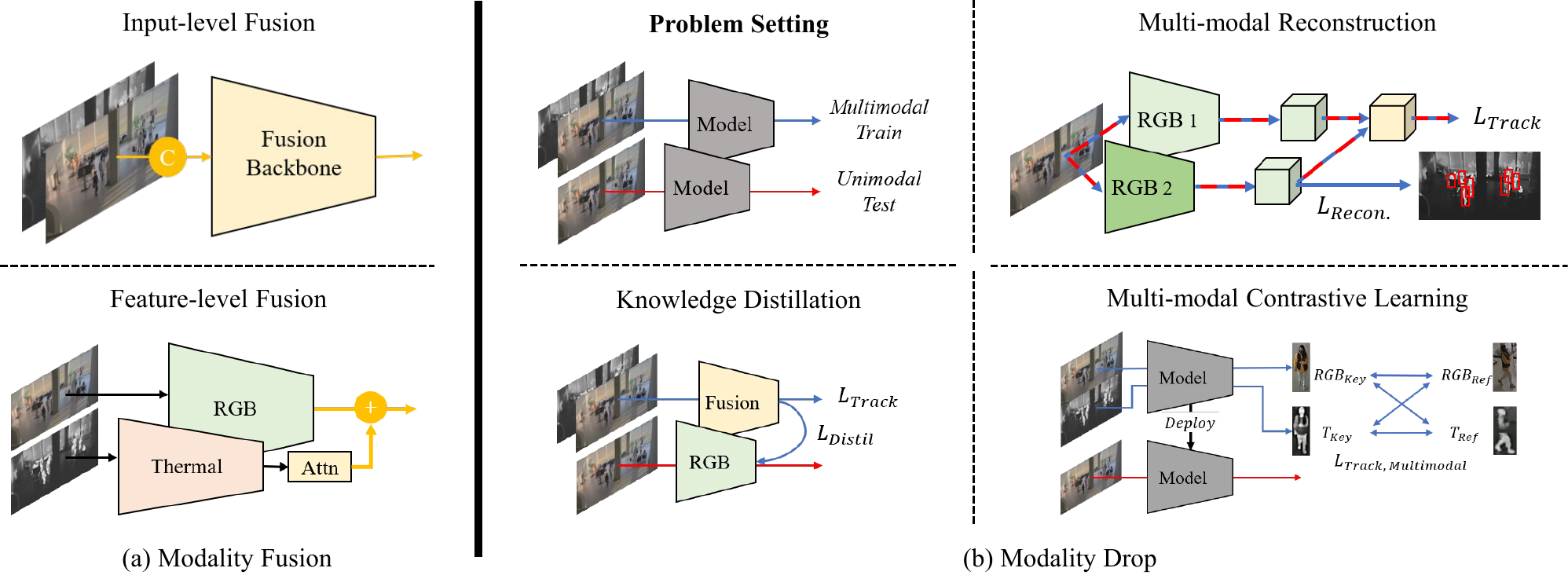}
  \vspace{-6mm}
  \captionsetup{font=footnotesize}
  \caption{\textbf{Multi-modal Learning Setups and Baselines.
  }(a) presents the concept of modality fusion with both input-level and feature-level fusion techniques integrating thermal data with RGB for enhanced object tracking. 
  (b) outlines the modality drop scenario, where the model trained on combined RGB and thermal data is tested solely on RGB data, using methods like multi-modal reconstruction, knowledge distillation, and multi-modal contrastive learning.
  }
  \vspace{-3mm}
  \label{fig:cross_modal_learning}
\end{figure*}

\subsection{Pre-Training: Real-world vs. Synthetic Data} \label{subsec:transfer}
In this study, we evaluate the efficacy of real-world data in improving MOT models by employing our MTMMC dataset as a foundational training set. We utilized the QDTrack \cite{pang2021quasi} as our base tracker and conducted experiments to measure its performance on the MOT17 benchmark. 
These experiments involved pre-training the model on the MTMMC dataset and subsequently fine-tuning it on MOT17. Additionally, we drew comparisons with models pre-trained on the MOTSynth dataset \cite{fabbri2021motsynth}, which is a large-scale synthetic dataset derived from extensive simulation within a gaming environment.

As detailed in Table \ref{tab:mot_synth}, our findings illustrate that the MTMMC dataset, albeit comprising half the number of annotations compared to MOTSynth (0.5M vs. 1M), and without the aid of complex data simulation techniques, still substantially contributes to the tracking accuracy. 
Notably, models pre-trained on MTMMC yield a MOTA score of 55.3 without fine-tuning (54.1 when pre-trained on MOTSynth) and see an increase to 68.6 with fine-tuning (70.8 when pre-trained on MOTSynth). While MOTSynth commences at a higher baseline, our real-world data, when combined with MOTSynth, demonstrates a remarkable synergy, resulting in a superior IDF1 score of 72.0 post fine-tuning.

These observations underscore the continued relevance of real-world datasets. 
While the scalability and control offered by synthetic data are appealing, the inherent complexities and variability present in real-world data are crucial for models to learn effectively. 
The MTMMC dataset, therefore, remains an invaluable resource for achieving high-fidelity tracking performance, and its integration with synthetic data further enhances this advancement.

\begin{table*}[t]
\setlength{\tabcolsep}{18pt}
 \centering
    \subfloat[\scriptsize Modality Fusion in MTMMC]
         {
         \resizebox{0.45\textwidth}{!}
        {
        \def\arraystretch{1.65}
        \begin{tabular}{l|c|ccc}
        \hline

        Method & Fusion & IDF1 & MOTA & mAP \\
        \hline
        RGB     & \xmark & 53.0 & 84.5 & 92.8 \\
        T       & \xmark & 44.5 & 79.2 & 89.9 \\
        \hline
        RGBT-I  & Input & \textbf{54.0} & 85.6 & 93.1 \\
        RGBT-F  & Feature & 53.9 & \textbf{86.0} & \textbf{93.5} \\
        \hline
        \end{tabular}
        }
         
         }
    \subfloat[\scriptsize Modality Drop in MTMMC $\rightarrow$ MOT17]
         {
         \resizebox{0.55\textwidth}{!}
        {
        \def\arraystretch{1.38}
        \begin{tabular}{l|cc|cc}
        \hline
        \multicolumn{1}{l|}{\multirow{2}{*}{Method}} & \multicolumn{2}{c|}{w/o fine-tune} & \multicolumn{2}{c}{w/ fine-tune} \\ \cline{2-5}
                                                     & IDF1 & MOTA & IDF1 & MOTA \\
        \hline
        RGB-Unimodal (baseline) & 55.3 & 43.6 & 68.6 & 66.6 \\
        \hline
        Knowledge Distill. & 55.1 & 43.2 & \textbf{70.5} & \textbf{68.0} \\
        Multi-modal Recon. & 57.9 & 46.2 & 68.3 & 67.6 \\
        Multi-modal Contrastive. & \textbf{59.7} & \textbf{48.4} & 68.3 & 67.3 \\
        \hline
        \end{tabular}
        }
    }
    \vspace{-3mm}
    \captionsetup{font=footnotesize}
    \caption{\textbf{Multi-modal learning results.}}
    \vspace{-2mm}
\label{table:rgbt_mot}
\end{table*}

\begin{table*}[!t]
    \setlength{\tabcolsep}{20pt}
    \centering
    \subfloat[\scriptsize RGB-based MTMC]
         {
         \resizebox{0.6\textwidth}{!}
         {
         \def\arraystretch{1.5}
         \begin{tabular}{c|ccccc}
             \hline
             Method & IDF1 & MOTA & FP & FN & IDs \\ 
             \hline
             TrackTA    & 32.8 &76.9 &10604 & 18715 & 13364 \\
             H. Cluster & 41.6 & 80.9 & 8012 & 14663 & 11072 \\ 
             \hline
        \end{tabular}
    }
    }
    \subfloat[\scriptsize Multi-modal MTMC]
        {
         \resizebox{0.4\textwidth}{!}
         {
         \def\arraystretch{1.6}\setlength{\tabcolsep}{11pt}
         \begin{tabular}{c|ccccc}
             \hline
             Fusion & IDF1 & MOTA & FP & FN & IDs \\ 
             \hline
             RGBT-I   &42.2 & 81.1 & 7823 & 14264 & 10803 \\
             RGBT-F   &43.5 & 81.7 & 7301 & 13592 & 9916 \\ 
             \hline
        \end{tabular}
    }
    }
    \vspace{-2mm}
    \captionsetup{font=footnotesize}
    \caption{\textbf{Multi-Target Multi-Camera Tracking Results} in MTMMC.
    For the efficient evaluation, we temporally sub-sampled the videos in 1FPS.
    H. Cluster denotes hierarchical clustering.
    The averaged results of all the testing scenarios are shown.}
    \label{tab:mtmc}
    \vspace{-3mm}
\end{table*}

\subsection{Multi-modal Learning: Setups and Baselines}

Multi-modal learning aims to improve scene understanding by leveraging complementary information from different sensor modalities. In this context, we explore how thermal data, when paired with RGB data, can enhance object tracking. This question stems from existing literature that demonstrates the benefits of such combinations in other domains~\cite{sun2019rtfnet,zhou2021gmnet,batchuluun2019action}. 
Our research extends these concepts into tracking scenarios using QDTrack~\cite{pang2021quasi} as the base tracker. 
We present two new learning setups, modality fusion and drop, illustrated in \figref{fig:cross_modal_learning}-(a) and (b), respectively.
We provide more detailed setup specifications and additional analyses in the appendix.
Here, we briefly introduce the high-level concepts of the setups and then discuss the key results.

\vspace{-2mm}
\paragraph{Modality Fusion}
We begin with modality fusion, focusing on the explicit integration of thermal data into RGB-based tracking models. This involves comparing both \emph{input} and \emph{feature-level} fusion methods against RGB and thermal-only baselines. We evaluate the benefits of thermal data incorporation, when it is directly available for both train and test.

\vspace{-2mm}
\paragraph{Modality Drop}
The modality drop setup presents a more challenging scenario. Here, the model is trained on both RGB and thermal data but is evaluated solely on RGB data. The rationale is that, during training, the model can learn generalized feature representations that are robust even when a modality is absent during testing. 
We introduce three strategies to harness RGB-T data effectively during training: \emph{knowledge distillation}, \emph{multi-modal reconstruction}, and \emph{multi-modal contrastive learning}. 

One practical application is using a multimodally trained tracking model in an unimodal tracking system.
For instance, consider CCTV surveillance systems, which predominantly rely on RGB cameras often due to hardware or budget constraints.
Our goal is to train the model using datasets like MTMMC, which contain both RGB and thermal data, and then test its effectiveness in environments that only provide RGB data.
Essentially, we aim to determine if the model can learn generic features from the combined RGB and thermal data during training, and preserve its tracking capabilities in the absence of thermal data during testing.




\paragraph{Results}
The results in \tabref{table:rgbt_mot}-(a) showcase the performance gains from modality fusion. The integration of thermal data at both the input (RGBT-I) and feature level (RGBT-F) with the base RGB data results in improved performance, compared to using RGB or thermal data in isolation. Notably, the RGBT-F approach, achieves the highest overall performance, with an IDF1 score of 53.9 and MOTA of 86.0.
This suggests that thermal data, when fused at the feature level, provides a more discriminative tracking representation.

In \tabref{table:rgbt_mot}-(b), we summarize the performance in the modality drop setup.
We simulate the modality drop scenario, by training the model using both RGB and thermal data in the MTMMC, and evaluate or optionally fine-tune the model using MOT17, which only provides RGB data.
Here, the `without fine-tuning' demonstrates how well the features learned from the combined multimodal data (RGB+T) transfer directly to the RGB domain.
On the other hand,
`with fine-tuning' evaluates how effectively these learned features serve as initialization for further refinement.
Without fine-tuning, Knowledge Distillation (KD) lags in performance (IDF1: 55.1, MOTA: 43.2), which is likely due to its strong dependence on thermal data imposed during distillation, resulting in a weaker generalization ability. In contrast, the Multi-modal Contrastive method shows a relatively high resilience (IDF1: 59.7, MOTA: 48.4), suggesting it learns modality-invariant features through contrastive learning, which confers strong generalization. 
With fine-tuning, KD exhibits a marked improvement (IDF1: 70.5, MOTA: 68.0), indicating its potential once adapted to the RGB domain. Conversely, the Multi-modal Contrastive method sees only a marginal increase after fine-tuning (IDF1: 68.3, MOTA: 67.3).
It is important to note that generalizable features do not necessarily equate to an optimal initialization for RGB-specific fine-tuning. 
We recognize the further investigations are necessary to fully understand the underlying mechanisms, and we leave this for future studies.

\subsection{Multi-modal MTMC}\label{subsec:mtmc}
Multi-target multi-camera (MTMC) expands upon MOT by requiring the identification of multiple targets across various camera views.
We build a strong baseline model to benchmark the MTMC scores on our new MTMMC dataset.
Specifically, we integrate the multi-object tracker and person Re-ID networks, QDtrack~\cite{pang2021quasi} and BoT~\cite{luo2019bag} to generate the tracklet-level feature representation.
Upon this, we study the performance of two leading multi-camera association (MCA) methods, optimization-based~\cite{he2020multi} and clustering-based~\cite{kohl2020mta}.

In~\tabref{tab:mtmc}-(a), the results show that the hierarchical clustering-based MCA method~\cite{kohl2020mta} outperformed the optimization-based approach~\cite{he2020multi}, which required heavy hyper-parameter tuning.
Table~\ref{tab:mtmc}-(b) presents the results following the integration of thermal information on the clustering-based method. 
The feature-fusion approach again resulted in more accurate multi-camera tracking.
As a dataset and evaluation paper, we focus on establishing baseline models and benchmark scores to set a stage for followup researches. We hope to see numerous advanced multi-modal tracking models presented upon our results.


\section{Conclusion}
We have presented the MTMMC dataset—a large-scale, real-world, multi-modal tracking benchmark designed to advance MTMC tracking. Through our extensive experiments, we have demonstrated its efficacy in improving the performance of various sub-tasks and have highlighted its synergistic use with synthetic data for pre-training. Additionally, we introduced two new multi-modal learning setups—modality fusion and drop—and developed robust baseline models for multi-modal MTMC tracking. We hope that our contributions will reinvigorate research in MTMC and will spark new innovations in multi-modal tracking technologies, ushering in a new era of intelligent tracking systems.

\noindent\textbf{Acknowledgement}
This work was partially supported by the NRF (NRF-2020M3H8A1115028, FY2021).

\appendix
\addcontentsline{toc}{section}{Appendices}
\vspace{10mm}

\twocolumn[{%
\section{Appendix}
\vspace{10mm}
\renewcommand\twocolumn[1][]{#1}%
\begin{center}
    \centering
    \includegraphics[width=\textwidth]{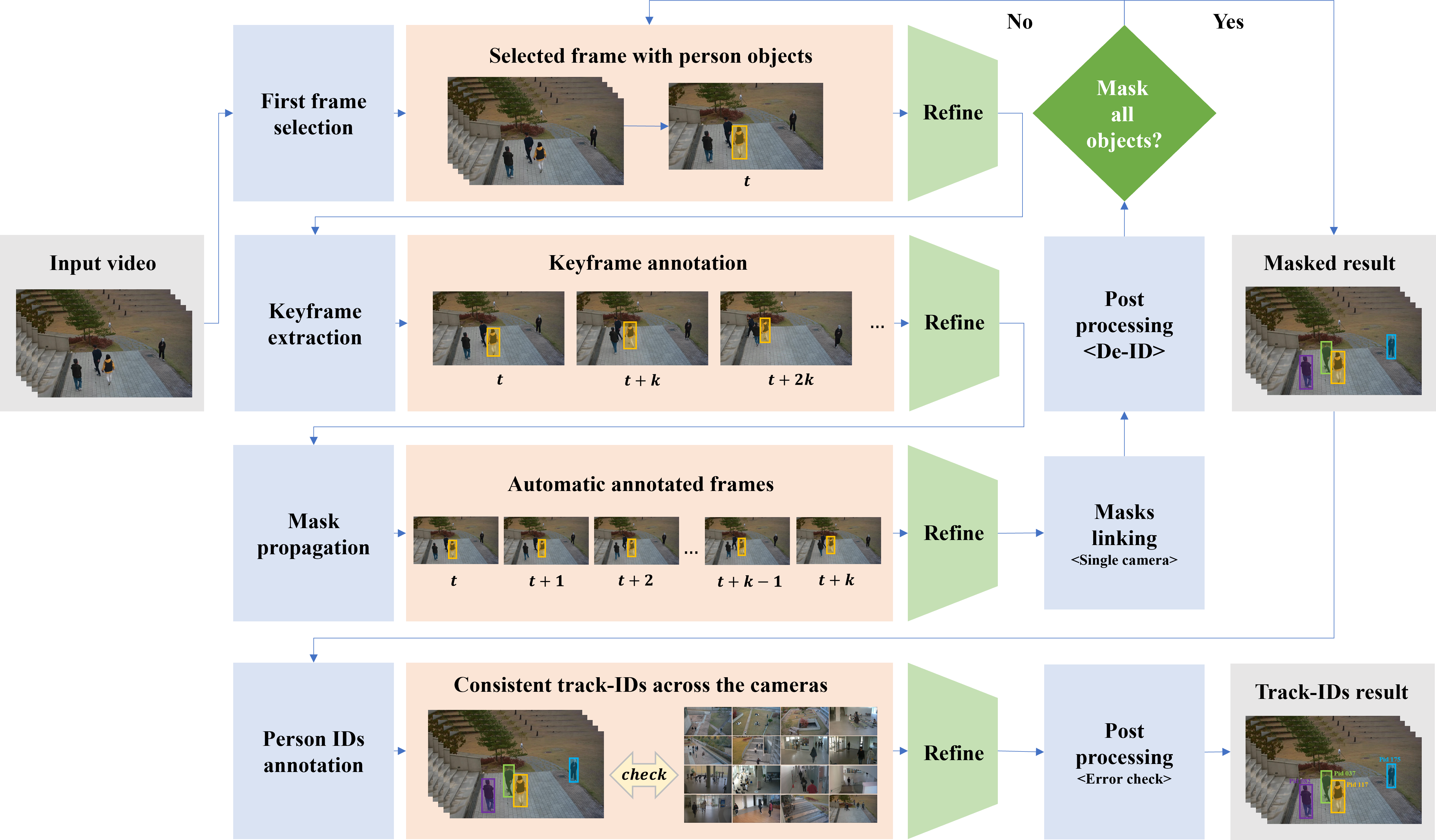}
    \vspace{3mm}
      \captionsetup{font=footnotesize}
      \captionof{figure}{\textbf{Single-camera tracking annotation pipeline.} We adopt the semi-automatic labeling approach. The workers first label the key frames and then the annotations for the other frames are interpolated based on the model predictions.}
    \label{fig:Annotation_pipeline}
\end{center}%
}]

\vspace{20mm}

\noindent In this supplementary material, we present detailed information on the following aspects:
\begin{enumerate}[label=\Alph*), topsep=-1pt, itemsep=1pt]
    \item Details of Annotation,
    \item Experimental Setup Specifications,
    \item Supplementary Experiments,
    \item Specifications of Camera Hardware,
    \item Overview of the MTMMC Dataset,
    \item Licenses of the Datasets Used,
    \item Video Demonstration, and
    \item Discussion of Ethical Considerations.
\end{enumerate}

\vspace{20mm}
\section{Annotation Details}
\subsection{Annotation Pipeline}


Our annotation pipeline is illustrated in \figref{fig:Annotation_pipeline}.
We separate the single-camera tracking task from the multi-camera association.
The annotation tool is built upon the CVAT~\footnote{https://github.com/maheriya/cvat}, an open-source vision annotation tool.
We further enabled several functionalities such as uploading large-scale videos, efficient task management of crowd workers, and text description translations.

\clearpage
\begin{figure*}[!h]
  \includegraphics[width=\linewidth]{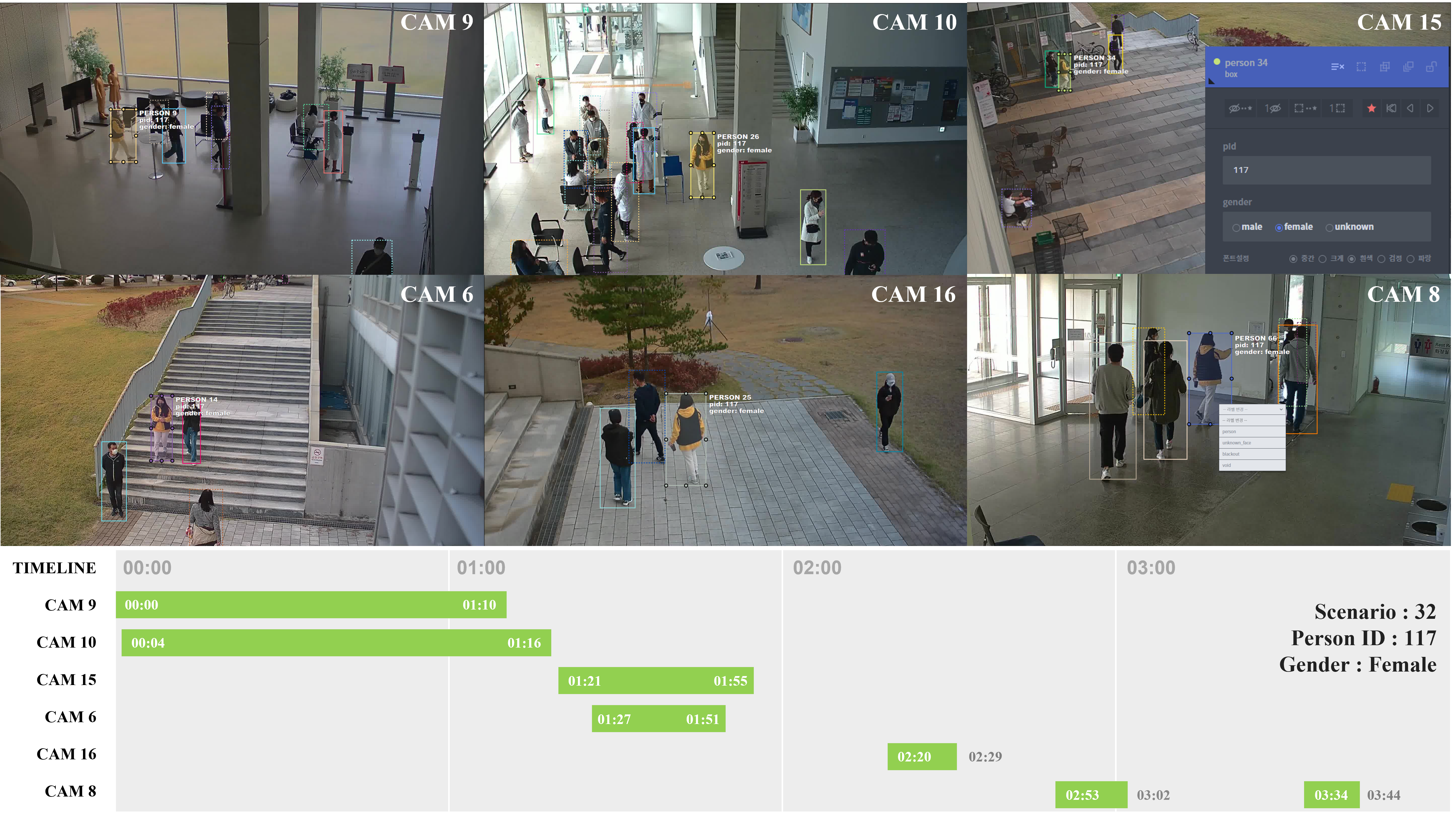}
  \vspace{-4mm}
  \captionsetup{font=footnotesize}
  \caption{\textbf{Multi-camera association.} The workers are instructed to assign consistent PIDs for the same person across the cameras for each scenario.}
  \vspace{-6mm}
  \label{fig:PIDtrack}
\end{figure*}
\begin{figure*}
    \centering
    \subfloat[]
    {{\includegraphics[width=0.19\textwidth]{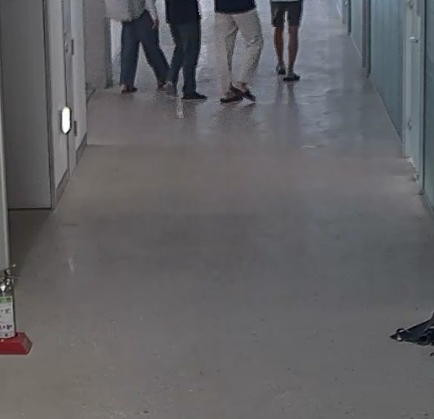}}}
    \hfill
    \subfloat[]
    {{\includegraphics[width=0.19\textwidth]{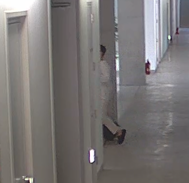}}}
    \hfill
    \subfloat[]
    {{\includegraphics[width=0.19\textwidth]{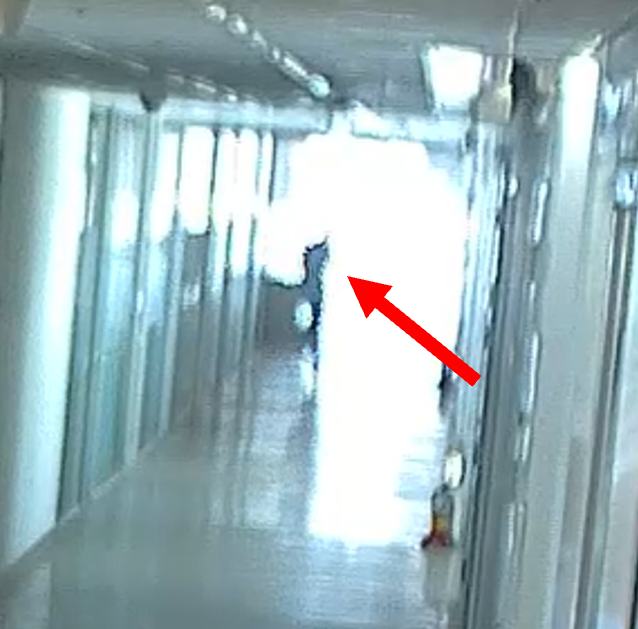}}}
    \hfill
    \subfloat[]
    {{\includegraphics[width=0.19\textwidth]{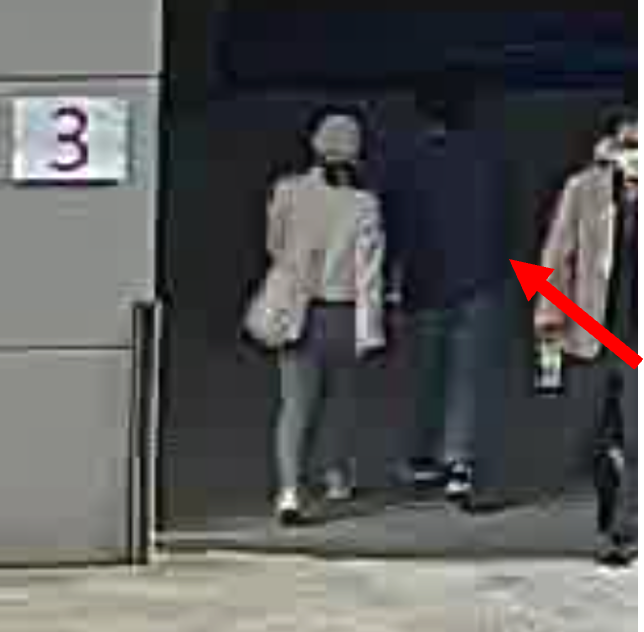}}}
    \hfill
    \subfloat[]
    {{\includegraphics[width=0.19\textwidth]{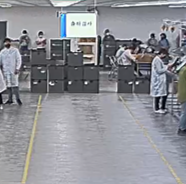}}}
    \vspace{-2.5mm}
    \captionsetup{font=footnotesize}
    \caption{\textbf{When to assign dummy person IDs.} (a),(b) horizontal and vertical truncation (c),(d) severe lighting (e) small-scale.}
    \vspace{-3.5mm}
    \label{fig:unidentifiablePID}
\end{figure*}
\begin{figure*}
    \centering
    \subfloat[]
    {{\includegraphics[width=0.24\textwidth]{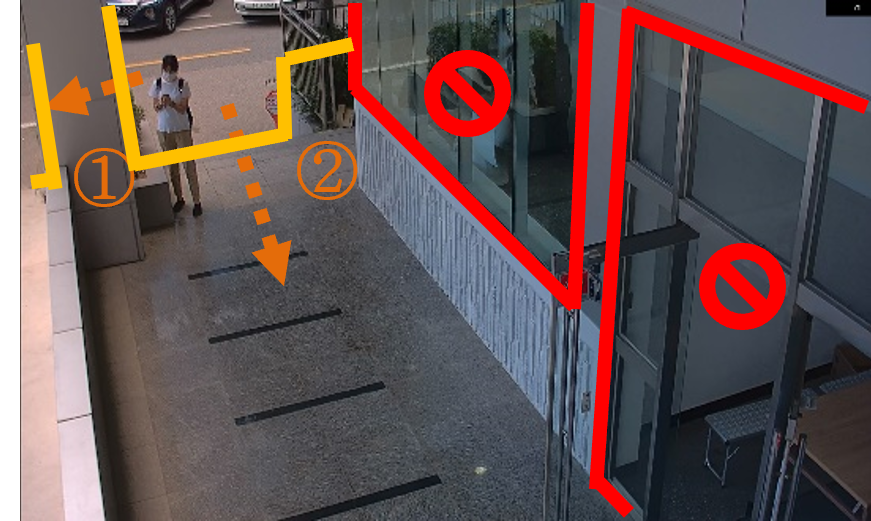}}}
    \hfill
    \subfloat[]
    {{\includegraphics[width=0.24\textwidth]{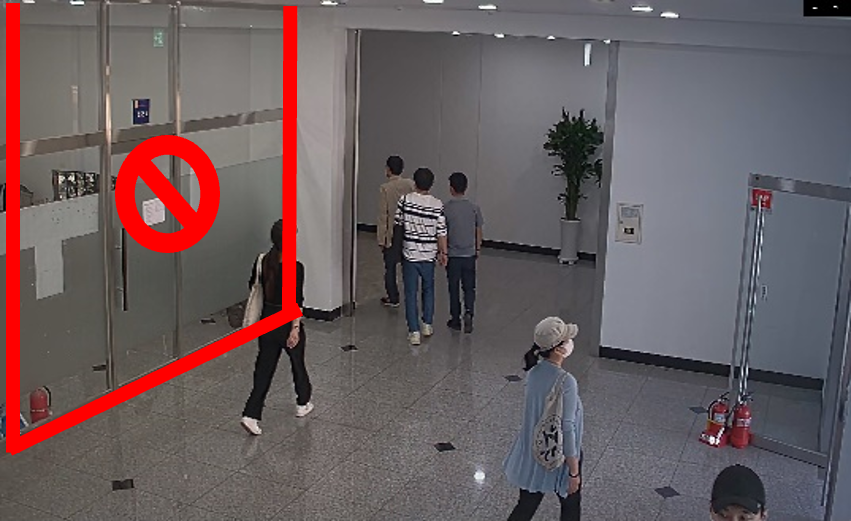}}}
    \hfill
    \subfloat[]
    {{\includegraphics[width=0.24\textwidth]{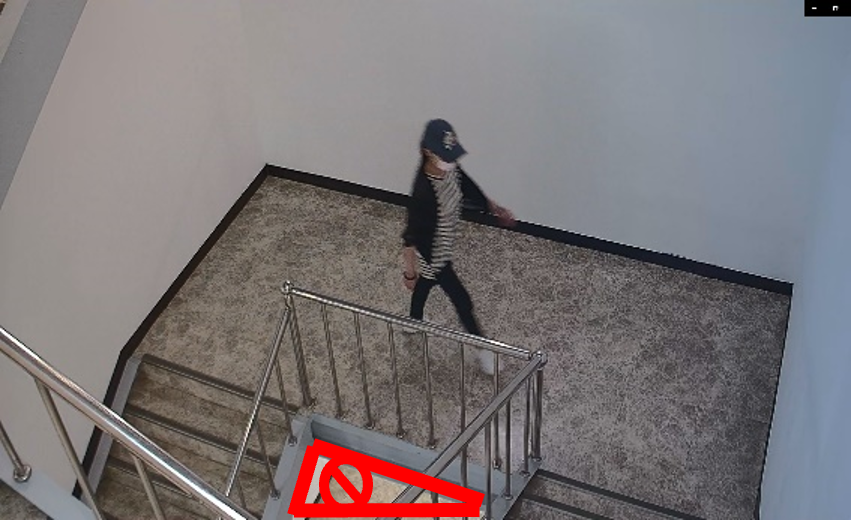}}}
    \hfill
    \subfloat[]
    {{\includegraphics[width=0.24\textwidth]{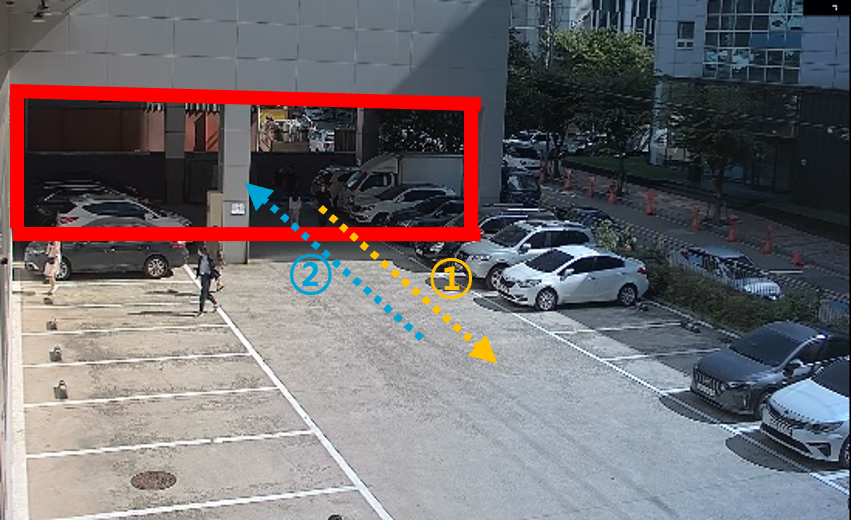}}}
    \vspace{-2.5mm}
    \captionsetup{font=footnotesize}
    \caption{\textbf{When to ignore the labeling.} (a),(b) reflection on transparent objects (c) severe truncation (d) poor lighting.}
    \vspace{-3.5mm}
    \label{fig:annotationscope}
\end{figure*}
\begin{figure*}
    \centering
    \subfloat[]
    {{\includegraphics[width=0.24\textwidth]{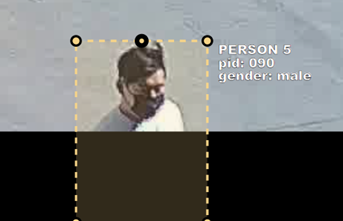}}}
    \hfill
    \subfloat[]
    {{\includegraphics[width=0.24\textwidth]{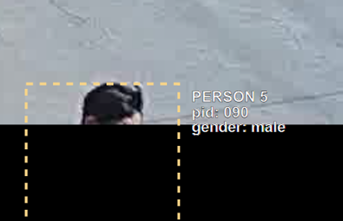}}}
    \hfill
    \subfloat[]
    {{\includegraphics[width=0.24\textwidth]{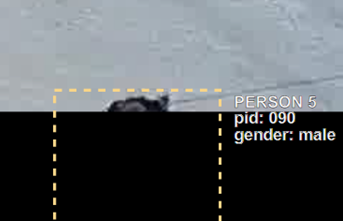}}}
    \hfill
    \subfloat[]
    {{\includegraphics[width=0.24\textwidth]{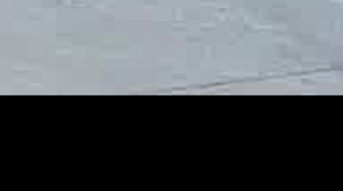}}}
    \vspace{-2.5mm}
    \captionsetup{font=footnotesize}
    \caption{\textbf{Amodal annotation example.}}
    \label{fig:frameborder}
\end{figure*}
\clearpage

\noindent\textbf{Single-camera tracking}
We employed 189 workers, 12 reviewers, and 2 project managers for our annotation task.
The project managers provided annotation guidelines and managed the scheduling.
As detailed in our main paper, we adopted a semi-automatic labeling approach.
We manually annotated keyframes (with a default temporal stride of 5 frames) and used the deepSORT algorithm~\cite{wojke2017simple} for interpolating annotations on intermediate frames.
Workers had the flexibility to adjust the temporal stride.
To ensure high-quality annotations, we conducted rigorous peer reviews. In instances where bystanders were captured, we applied face blurring in post-processing to address privacy concerns. 
On average, each worker annotated 525 images per day. All annotations were saved in JSON format.

\vspace{3mm}
\noindent\textbf{Multi-camera Association}
For this task, we selected a team of 8 highly skilled workers, supported by 8 reviewers and 2 project managers. Initially, we attempted model prediction-based labeling, but it failed to meet our internal quality standards.
Consequently, as illustrated in \figref{fig:PIDtrack},  we shifted to manual labeling. Workers used the labels generated from the single-camera tracking phase as a base, checking for errors and matching labels across different cameras to assign final Person IDs (PIDs). In situations with ambiguous identifications, workers were permitted to use a placeholder label of `0'. To guarantee the accuracy of our data, we enforced two essential camera constraints and iteratively corrected any discrepancies until we observed no significant errors.

\subsection{Annotation Instructions}
To ensure high-quality annotations, we established specific guidelines for workers to handle exceptional cases.
These guidelines were designed to maintain consistency and accuracy in challenging annotation scenarios.
\begin{itemize}[leftmargin=*]\itemsep0.5em
    \item \textbf{Dummy Person ID} 
    In instances where individuals were challenging to identify due to significant changes in appearance, scale, or lighting, workers assigned a placeholder ID of '0'. This practice was crucial for maintaining data integrity in cases where person identification was ambiguous or unreliable (refer to \figref{fig:unidentifiablePID}).
    \item \textbf{Ignore Area} 
    Our guidelines specified that reflections of persons on transparent surfaces, such as glass doors, should be excluded from annotations. Similarly, extreme cases of body part truncation or poor lighting conditions were treated as grey areas and omitted from the dataset to ensure annotation quality (refer to \figref{fig:annotationscope}).
    \item \textbf{Amodal Annotation} 
    Following the MOT17 annotation standard \cite{milan2016mot16}, we instructed workers to label objects amodally. This approach involved marking occlusions distinctly from visible parts using dotted lines, enhancing the precision of our dataset (refer to \figref{fig:frameborder}).
\end{itemize}





\section{Experimental Details}
\subsection{Datasets}
We provide the details of the additional data source used for the experiments in the main paper.
\\
\\
\noindent\textbf{COCO2017} 
COCO~\cite{lin2014microsoft} includes object detection and segmentation labels with 118k training images and 5k validation images.
We collected only the person detection labels, COCO-Person, and used them for the person detection experiments.
\\
\\
\noindent\textbf{Market-1501} 
Market-1501~\cite{zheng2015scalable} is a popular person re-ID dataset collected from six outdoor cameras.
It contains 32,668 bounding boxes of 1,501 identities.
To simulate the realistic scenarios, Deformable Part Model (DPM)~\cite{felzenszwalb2010object} is employed to produce bounding boxes of pedestrians. We used it for a person re-identification experiments.
\\
\\
\noindent\textbf{MSMT17}
MSMT17~\cite{wei2018person} is a challenging person re-identification dataset with 126,441 bounding boxes of 4,101 identities captured by 15 outdoor cameras and predicted using Faster RCNN~\cite{ren2015faster}. The dataset includes complex scenes and significant variations in lighting and viewpoints to provide challenging cases. We utilized MSMT17 for our person re-identification experiments.
\\
\\
\noindent\textbf{MOT17} 
MOT17~\cite{milan2016mot16} is a widely used multi-object tracking benchmark that includes 7 training and 7 testing videos with challenging scenarios, such as frequent occlusions and crowd scenes. We split each training sequence into two halves and used the first half-frames for training and the second half for validation. We utilized MOT17 for single-camera tracking experiments.
\\
\\
\noindent\textbf{MOTSynth}
MOTSynth~\cite{fabbri2021motsynth} is a large-scale synthetic dataset created using the photorealistic video game Grand Theft Auto V. It is designed for person tracking and segmentation in urban scenarios and contains 764 full-HD videos, 1.3M frames, and 33M person instances. We utilized the official train and validation split and applied it to our transfer learning experiments.

\subsection{Implementation Details}

\paragraph{Training}
Our framework is constructed using two well-established public codebases: \textit{mmtracking}~\cite{mmtrack2020} and \textit{fast-reid}~\cite{he2020fastreid}.
We adhere closely to the default training recipes provided by these codebases, including data augmentation techniques and training schedules detailed in \tabref{tab:train_Recipes}. 
For implementation, we utilized Pytorch v1.10 and CUDA v11.3, executing our models on a robust hardware setup equipped with an AMD EPYC 7352 (2.3GHz) CPU and an NVIDIA RTX A6000 GPU, ensuring efficient processing and analysis.


\begin{table*}[t!]
\setlength{\tabcolsep}{12pt}
\centering
\resizebox{\linewidth}{!}{
\begin{tabular}{c|c|llllll}
\textbf{Sub-task}                          & \textbf{Model}        & \multicolumn{1}{c}{\textbf{epoch}} & \multicolumn{1}{c}{\textbf{fine-tune}} & \multicolumn{1}{c}{\textbf{optimizer}} & \multicolumn{1}{c}{\textbf{lr}} & \multicolumn{1}{c}{\textbf{momentum}} & \multicolumn{1}{c}{\textbf{weight\_decay}} \\ \hline
Detection & Faster R-CNN & \multicolumn{1}{c}{24}             & \multicolumn{1}{c}{4}                  & \multicolumn{1}{c}{SGD}                & \multicolumn{1}{c}{0.02}        & \multicolumn{1}{c}{0.9}               & \multicolumn{1}{c}{0.0001}                 \\ \hline
                                           & AGW          &     \multicolumn{1}{c}{120}                               &        \multicolumn{1}{c}{-}                                &          \multicolumn{1}{c}{adam}                                                      &           \multicolumn{1}{c}{0.00035}                      &        \multicolumn{1}{c}{-}                               &                \multicolumn{1}{c}{0.0005}                            \\ 
\multirow{-2}{*}{Re-ID}           & BOT          &     \multicolumn{1}{c}{120}                         &           \multicolumn{1}{c}{-}                             &                  \multicolumn{1}{c}{adam}                                              &      \multicolumn{1}{c}{0.00035}                           &                 \multicolumn{1}{c}{-}                      &                \multicolumn{1}{c}{0.0005}                            \\ \hline
                                           & JDE          & \multicolumn{1}{c}{30}                                    &  \multicolumn{1}{c}{30}                                       &  \multicolumn{1}{c}{SGD}                                                              & \multicolumn{1}{c}{0.01}                                & \multicolumn{1}{c}{0.9}                                      & \multicolumn{1}{c}{0.0001}                                           \\
                                           & QDTrack      & \multicolumn{1}{c}{4}              & \multicolumn{1}{c}{4}                  & \multicolumn{1}{c}{SGD}                & \multicolumn{1}{c}{0.02}        & \multicolumn{1}{c}{0.9}               & \multicolumn{1}{c}{0.0001}                 \\
                                           & CenterTrack  & \multicolumn{1}{c}{70}             & \multicolumn{1}{c}{best}               & \multicolumn{1}{c}{adam}                                       & \multicolumn{1}{c}{0.0125}      & \multicolumn{1}{c}{-}               & \multicolumn{1}{c}{0.0001}                 \\
\multirow{-4}{*}{MOT}             & ByteTrack    & \multicolumn{1}{c}{300}            & \multicolumn{1}{c}{40}                 & \multicolumn{1}{c}{SGD}                & \multicolumn{1}{c}{0.01}        & \multicolumn{1}{c}{0.9}               & \multicolumn{1}{c}{0.0005}                 \\  
\end{tabular}

}
\captionsetup{font=footnotesize}
\caption{\textbf{Subtasks Training Recipes}}
\label{tab:train_Recipes}
\end{table*}


\begin{table*}[t]
\setlength{\tabcolsep}{8pt}
 \centering
    \subfloat[\scriptsize Feature-level Fusion]
         {
         \resizebox{0.50\textwidth}{!}
        {
        \def\arraystretch{1.65}
        \begin{tabular}{l|ccc|ccc}
        \hline

        Method & In. & Attn. & Asy. & IDF1 & MOTA & mAP \\
        \hline
        RGB     &- & - & - & 53.0 & 84.5 & 92.8 \\
        \hline
        \multirow{1}{*}{(1)}        & Add. & BAM & \checkmark & 53.7 & 85.3 & 93.1 \\
        \hline
        \multirow{2}{*}{(2)}        & Diff. & Cha. & \checkmark & 53.5 & 85.1 & 93.0 \\
                & Diff. & Spa. & \checkmark & 53.7 & 85.4 & 93.3 \\
        \hline
        \multirow{1}{*}{(3)}        & Diff. & BAM & \xmark & 53.8 & 85.6 & 93.2 \\
        \hline
        RGBT-F  & Diff. & BAM & \checkmark & \textbf{53.9} & \textbf{86.0} & \textbf{93.5} \\
        \hline
        \end{tabular}
        }
         
         }
    \subfloat[\scriptsize Multi-modal contrastive-learning]
         {
         \resizebox{0.5\textwidth}{!}
        {
        \def\arraystretch{1.6}
            
            \begin{tabular}{c|c|c|ccc}
            \hline
            Modality              & Network                   & \# contra.loss pair & IDF1             & MOTA             & mAP              \\ \hline
            RGB                   & baseline                  & 1                   & 55.7          & 43.8          & 55.6          \\ \hline
            \multirow{6}{*}{RGB-T} & \multirow{2}{*}{parallel} & 2                   & 56.1          & 44.5          & 57.5          \\
                                  &                           & 4                   & 56.8          & 44.3          & 57.6          \\ \cline{2-6} 
                                  & \multirow{4}{*}{share}    & 2                   & 55.6          & 43.5          & 56.7          \\
                                  &                           & 4                   & \textbf{59.7} & \textbf{48.4} & \textbf{59.7} \\
                                  &                           & 6                   & 57.3          & 44.5          & 57.3          \\
                                  &                           & 8                   & 57.5          & 45.4          & 58.1          \\ \hline
            \end{tabular}

        }
         
         }
    \vspace{-2mm}
    \captionsetup{font=footnotesize}
    \caption{\textbf{Ablation studies of multi-modal learning models on MTMMC.}}
    \vspace{-3mm}
\label{table:ablation_study}
\end{table*}

\paragraph{Multi-model Learning}
In this section, we detail the design of two proposed multi-modal learning setups.

\paragraph{1) Modality Fusion}
\label{paragraph:Modality_fusion}
\begin{itemize}[leftmargin=*]\itemsep0.5em
    \item \textbf{Input-level fusion.} 
    We employ channel-wise concatenation~\cite{eigen2015predicting} to combine RGB and Thermal inputs into a 4-channel RGB-T input. This involves modifying the first convolutional layer in the backbone to accept 4 channels instead of 3, initializing the additional channel's weight as the average of the RGB channel weights.
    
    \item \textbf{Feature-level fusion.} 
     Building upon prior research \cite{zhang2019cross,deng2019rfbnet,zhou2021mffenet}, we have developed a conditional attention module to utilize multi-modal data more effectively. This module uses a BAM attention block~\cite{park2018bam}, processing the differential between RGB and thermal features to explicitly highlight the complementary nature of these modalities. 
     The fusion process is executed as $F_{fuse}=F_{rgb}+BAM(F_{thermal} - F_{rgb}) \otimes F_{thermal}$, where $F_{fuse}$ represents the fused feature output and $\otimes$ is element-wise multiplication. We integrate this attention module at various levels within the Feature Pyramid Network (FPN~\cite{lin2017feature}), training it end-to-end without additional adjustments.
\end{itemize}
\vspace{1em}
To empirically validate our design choices, we conducted ablation studies as shown in Table \ref{table:ablation_study}-(a).
We investigated three important design choices: (1) input for the attention module, (2) attention design, and (3) asymmetric encoders.
Firstly, we found that forwarding the feature difference (Diff.) between the thermal and RGB performed better than their addition (Add.).
Secondly, while using either channel (Cha.) or spatial (Spa.) attention resulted in performance improvements over the simple baseline model, jointly using them (BAM) yielded the best performance, as noted in the original paper \cite{park2018bam}.
Finally, the asymmetric encoder design, which is intended to properly cope with the input information density, resulted in better tracking performance.

\begin{figure*}[!t]
  \includegraphics[width=\linewidth]{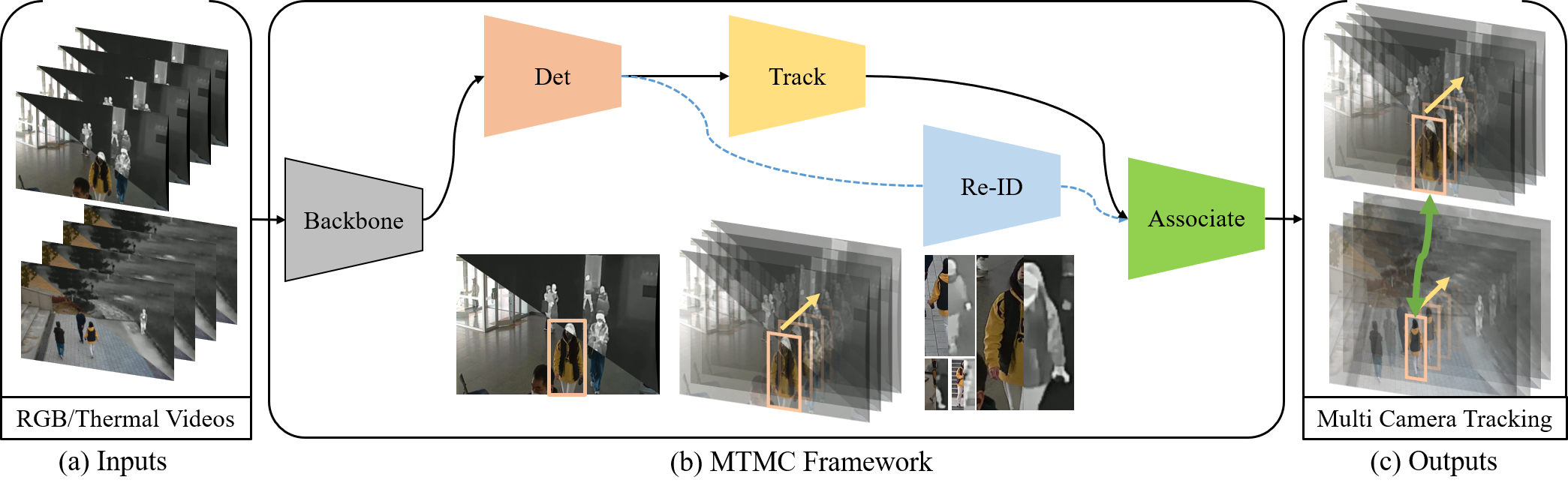}
  \vspace{-6mm}
  \captionsetup{font=footnotesize}
  \caption{\textbf{Overview of multi-target multi-camera tracker.}}
  \label{fig:network_overview}
\end{figure*}

\paragraph{2) Modality Drop}
\begin{itemize}[leftmargin=*]\itemsep0.5em
    \item \textbf{Knowledge distillation.} 
    To transfer the knowledge learned from the RGBT-F model to the standard RGB model, we distill the final level of FPN features to the RGB model. We implement the distillation loss using the MSE loss function, defined as $L_{Distill} = L_{MSE}(F_{rgb},F_{fuse})$. The final objective function is defined as $L = L_{Track} + \lambda L_{Distill}$, where $L_{Track}$ represents the total loss to train the base tracker~\cite{pang2021quasi}. The $\lambda$ is set to 0.1.
    
    \item \textbf{Multi-modal reconstruction.}
    Motivated by the MSDN framework~\cite{xu2017learning}, we incorporate the multi-modal reconstruction loss into the base tracker design. Specifically, we use two identical backbones, $B_{1}$ and $B_{2}$, and extract RoI features $F_{1}$ and $F_{2}$ from the RGB input. We then use $F_{2}$ to reconstruct the corresponding thermal information with a single deconvolution layer, which enforces $F_{2}$ to encode the RGB-to-Thermal correlations explicitly. The fused features of $F_{1}$ and $F_{2}$ are then forwarded to the tracking head, implicitly encoding the thermal information without its presence at test time. The total loss is $L = L_{track} + \lambda L_{recon}$, where $\lambda$ is set to 1.0.
    
    \item \textbf{Multi-modal contrastive-learning.}
    To enhance the multi-view contrastive learning, we made two adaptations to the multiple positive contrastive loss~\cite{pang2021quasi} that match the RGB instance features across the key and reference frames. Firstly, we allowed the instance feature matching within the frames, such as (key-key). Secondly, we included the instance thermal features, resulting in new feature combinations, such as RGB-T or T-T. The anchor features were selected from the key frame, and the positive and negative features were selected from the reference frame. 
   The following combinations are investigated:
   
    \begin{itemize}\footnotesize
        \item $L^{2-pair}_{contra} = RGB_{Key}-RGB_{Ref} + T_{Key}-T_{Ref}$ 
        \item $L^{4-pair}_{contra} = L^{2-pair}_{contra} + RGB_{Key}-T_{Ref} + T_{Key}-RGB_{Ref}$
        \item $L^{6-pair}_{contra} = L^{4-pair}_{contra} + RGB_{Key}-T_{Key} + T_{Key}-RGB_{Key}$
        \item $L^{8-pair}_{contra} = L^{6-pair}_{contra} + RGB_{Ref}-T_{Ref} + T_{Ref}-RGB_{Ref}$
    \end{itemize}
   \vspace{.5em}
    In~\tabref{table:ablation_study}-(b), we investigate the impact of multi-view contrastive learning along with the model design. Firstly, we find that multi-view matching generally outperforms the original RGB-based single-view matching. Secondly, sharing the backbone encoder for two different input sources leads to better results.
\end{itemize}





In our empirical evaluation, we demonstrated that our Modality Fusion method significantly enhances tracking performance. Additionally, the Modality Drop approach effectively encodes multi-modal correlations, improving tracker generalizability. Building on these findings, we designed a multi-modal Multi-Target Multi-Camera (MTMC) system, as detailed in the architectural overview in
\figref{fig:network_overview}.
Our system employs a two-stage approach. Initially, it generates tracklets for each video using a tracking-by-detection mechanism. These tracklets are then linked within each video through a Multi-Camera Association (MCA) process.
In the main paper, we investigate the two primary MCA approaches: optimization-based~\cite{he2020multi} and clustering-based methods~\cite{kohl2020mta}.
Moreover, we have validated the effectiveness of incorporating additional thermal information as a prior, which significantly enhances the tracker's capabilities.

\begin{table}{
\setlength{\tabcolsep}{12pt}
\centering
\resizebox{0.40\textwidth}{!}{
\def\arraystretch{1.3}
\begin{tabular}{c|c|cc}
\hline
Method & Train   & Rank 1 & mAP \\
\hline
\hline
\multicolumn{1}{c|}{\multirow{3}{*}{AGW}} & Market-1501  & 98.0 & 65.2  \\
& MSMT17  & 94.0 & 69.9  \\
& MTMMC-reID  & \textbf{98.0}  & \textbf{72.5}\\
\hline
\multicolumn{1}{c|}{\multirow{3}{*}{BOT}} & Market-1501   & 96.0& 64.4  \\
& MSMT17  & 94.0  & 68.6 \\
& MTMMC-reID & \textbf{98.0}  & \textbf{69.5}  \\
\hline
\end{tabular}
}
\vspace{-2mm}
\captionsetup{font=footnotesize}
\caption{\textbf{Cross-domain re-identification results.}}\label{table:reid}
}
\end{table} 

\vspace{5mm}

\begin{table}{
\centering
\setlength{\tabcolsep}{20pt}
\resizebox{0.46\textwidth}{!}{
\def\arraystretch{1.2}
\begin{tabular}{c|c|cc}
\hline
Method               & Modality     & Rank 1                                        & mAP                                \\ \hline\hline
\multirow{3}{*}{AGW} & RGB & 78.8                                          & 45.7                               \\ \cline{2-4} 
                     & RGBT-I       & 79.2  & 47.3 \\
                     & RGBT-F       & \textbf{80.7}                & \textbf{48.4}     \\ \hline
\multirow{3}{*}{BOT} & RGB & 74.4                                          & 42.7                               \\ \cline{2-4} 
                     & RGBT-I       & 75.1         & 44.1        \\
                     & RGBT-F       & \textbf{77.7}                & \textbf{44.6}                     
\end{tabular}
}
\vspace{-2mm}
\captionsetup{font=footnotesize}
\caption{\textbf{Multi-modal re-identification results.}}
\label{table:MTMMCreid}
}
\end{table} 

\vspace{5mm}
 
\begin{table}[t!]
\centering
\setlength{\tabcolsep}{1pt}
\resizebox{0.97\linewidth}{!}{
\def\arraystretch{1.2}
\begin{tabular}{c|cc|cc|cc}
\toprule
\multicolumn{1}{c|}{\multirow{2}{*}{Method}} & \multicolumn{2}{c|}{Train set} & \multicolumn{2}{c|}{Eval on MTMMC}  & \multicolumn{2}{c}{Eval on MOT17} \\ \cline{2-3}\cline{4-5} \cline{6-7} 
\multicolumn{1}{c|}{}                         & MTMMC & MOT17 & IDF1 & MOTA   & IDF1 & MOTA \\ \midrule

\multirow{2}{*}{BoT-SORT}                   & \checkmark       &   & 64.7 &  89.2 & 70.8 & 56.7     \\
                                            &  & \checkmark         & 40.7 & 58.5   & 78.1  & 75.0      \\  \hline
\multirow{2}{*}{OC-SORT}                   & \checkmark       &   & 63.4 & 88.5  & 68.8  & 55.3      \\
                                            &  & \checkmark         & 39.2 & 52.8   & 74.5 & 73.5      \\  \bottomrule

\end{tabular}
}
\vspace{-2mm}
 \captionsetup{font=footnotesize}
 \caption{\textbf{Multi Object Tracking Results using SOTA trackers.}}
 \vspace{-3mm}
\label{tab:MOT_recent}
\end{table}

\section{Additional Experiments}
\paragraph{Cross-domain Person Re-Identification} 
To illustrate the broad applicability of our re-ID representation, we conducted cross-domain experiments in~\tabref{table:reid} using using two Re-ID models, BOT~\cite{luo2019bag} and AGW~\cite{ye2021deep}. 
Trained on pedestrian datasets (Market-1501~\cite{zheng2015scalable}, MSMT17~\cite{wei2018person}, and MTMMC-reID) and tested in sports scenes~\cite{van2022deepsportradar}, the model using our dataset demonstrated superior performance, indicating that our dataset provides more generic representations.


\paragraph{Multi-modal Person Re-Identification}
To further validate the effectiveness of incorporating thermal modality, we conducted multi-modal re-identification experiments. We implemented two fusion approaches: input-level fusion (RGBT-I) and feature-level fusion (RGBT-F). Our findings indicate performance enhancements in both the BOT~\cite{luo2019bag} and AGW~\cite{ye2021deep} models, thereby reaffirming the efficacy of integrating thermal modality in this context.

\paragraph{Multi Object Tracking w. SOTA trackers} 
In ~\tabref{tab:MOT_recent},
we present the results of MOT experiments using two recent models~\cite{aharon2022bot, cao2023observation}. The results demonstrate that these algorithms exhibit lower performance when trained and tested on our MTMMC dataset compared to their performance on the MOT17 dataset. This indicates that our MTMMC dataset presents a richer array of complexities, emphasizing its significance in training and testing more advanced models. 

\begin{figure*}[t!]
\begin{center}
\includegraphics[width=0.8\linewidth]{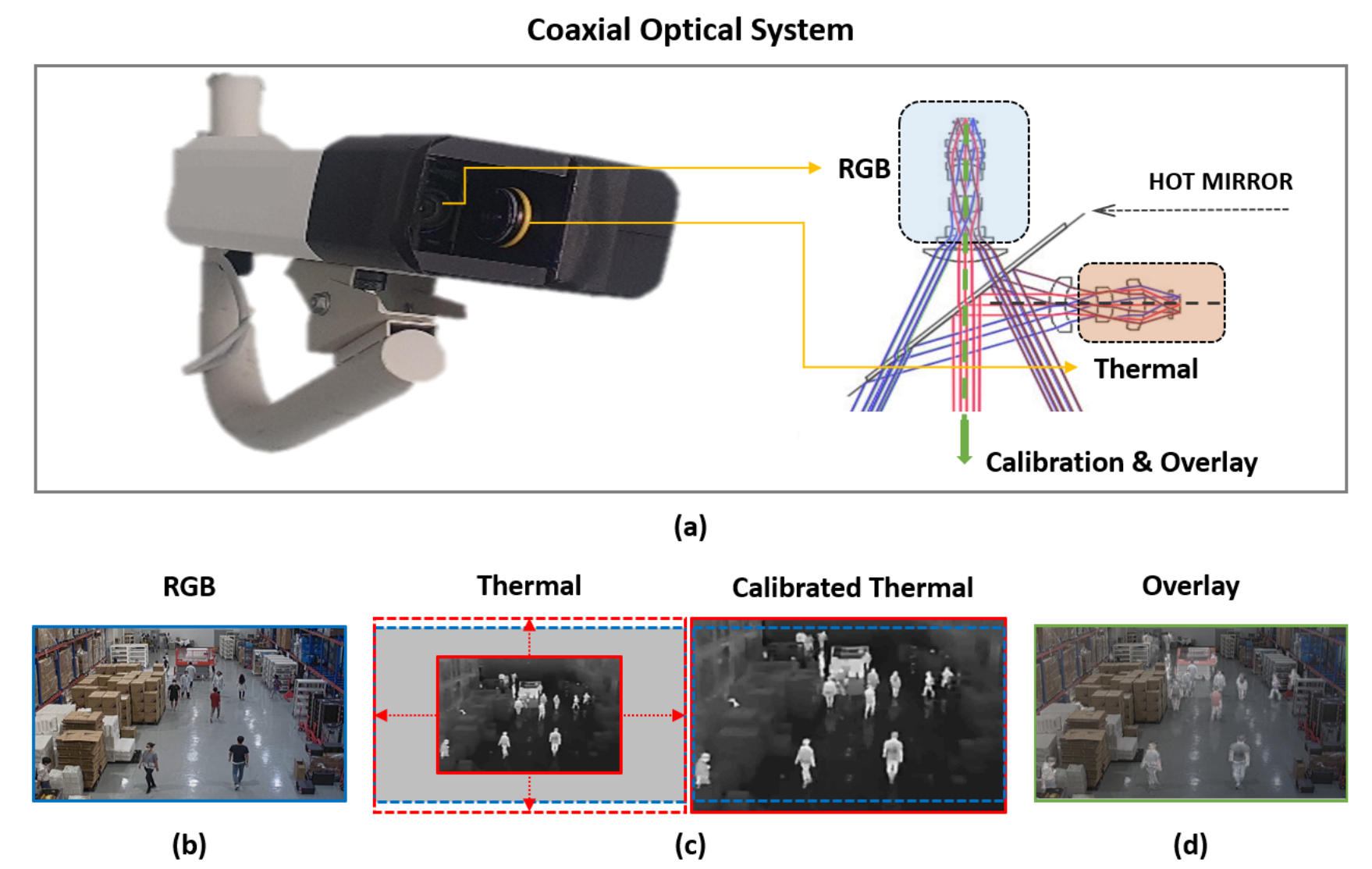}
\captionsetup{font=footnotesize}
\caption{(a) Multi-modal camera with Coaxial Optical System. (b) RGB image. (c) Calibrating Thermal image with RGB image. (d) Overlay image.} 
\label{fig:vision_structure}
\vspace{-3mm}
\end{center}
\end{figure*}

\section{Camera Hardware Specifics}


We detail our RGB-Thermal multi-modal camera system, adapted from Hwang et al.~\cite{hwang2015multispectral} for surveillance. It consists of an RGB camera (1920 × 1080 resolution) and a thermal camera (320 × 240 resolution), both with a 50$^{\circ}$ horizontal field of view and 60 fps frame rate. A hot mirror is included to filter extraneous radiation. For uniformity, thermal images are resized to 1920 × 1080, with border areas discarded during processing. See \figref{fig:vision_structure} (c) and (d) for further details.

\begin{table*}[!t]
\setlength{\tabcolsep}{6pt}
 \centering
    \subfloat[\scriptsize Environment]
         {
         \resizebox{0.3\textwidth}{!}
        {
        \def\arraystretch{1.42}
        \begin{tabular}{c|c|c|c}
        \hline
        \multirow{1}{*}{} & \multirow{1}{*}{Factory} & \multirow{1}{*}{Campus} & \multirow{1}{*}{Total} \\\hline
        trian (\#)                     & 7                               & 7                            & 14                     \\
        val (\#)                   & 3                               & 2                            & 5                      \\
        test (\#)                         & 2                               & 4                            & 6                      \\ \hline
        Total (\#)                        & 12                              & 13                           & 25                     \\ \hline
        \end{tabular}
        }
         
         }
    \subfloat[\scriptsize Weather]
         {
         \resizebox{0.3\textwidth}{!}
        {
        \def\arraystretch{1.35}
        \begin{tabular}{c|c|c|c}
        \hline
        \multirow{1}{*}{} & \multirow{1}{*}{Sunny} & \multirow{1}{*}{Cloudy} & \multirow{1}{*}{Total} \\\hline
        trian (\#)                     & 13                               & 1                            & 14                     \\
        val (\#)                   & 3                               & 2                            & 5                      \\
        test (\#)                         & 5                               & 1                            & 6                      \\ \hline
        Total (\#)                        & 21                              & 4                           & 25                     \\ \hline
        \end{tabular}
        }
         
         }
        \subfloat[\scriptsize Season]
         {
         \resizebox{0.3\textwidth}{!}
        {
        \def\arraystretch{1.3}
        \begin{tabular}{c|c|c|c}
        \hline
        \multirow{1}{*}{} & \multirow{1}{*}{Summer} & \multirow{1}{*}{Fall} & \multirow{1}{*}{Total} \\ \hline
        trian (\#)                     & 7                               & 7                            & 14                     \\
        val (\#)                   & 3                               & 2                            & 5                      \\
        test (\#)                         & 2                               & 4                            & 6                      \\ \hline
        Total (\#)                        & 12                              & 13                           & 25                     \\ \hline
        \end{tabular}
        }
         
         }
 \vspace{2mm}
 \captionsetup{font=footnotesize}     
 \caption{\textbf{Details on the dataset split.}}
\label{tab:data_split}
 \vspace{-3mm}
\end{table*}

\section{MTMMC Dataset Details}

\subsection{Dataset Split}
We divided the 25 MTMMC scenarios into three sets, consisting of 14, 5, and 6 scenarios for the training, validation, and testing sets, respectively. The division was based on the meta information to ensure equal and well-distributed representation, shown in Table \ref{tab:data_split}.

\subsection{{Dataset Statistics}}
This section presents a detailed analysis of our dataset through three key metrics: (1) objects per frame, (2) tracks per video, and (3) age and gender distribution. These metrics highlight the dataset's unique characteristics, enhancing its effectiveness for multiple object tracking tasks.

\begin{figure*}[!h]
  \includegraphics[width=\linewidth]{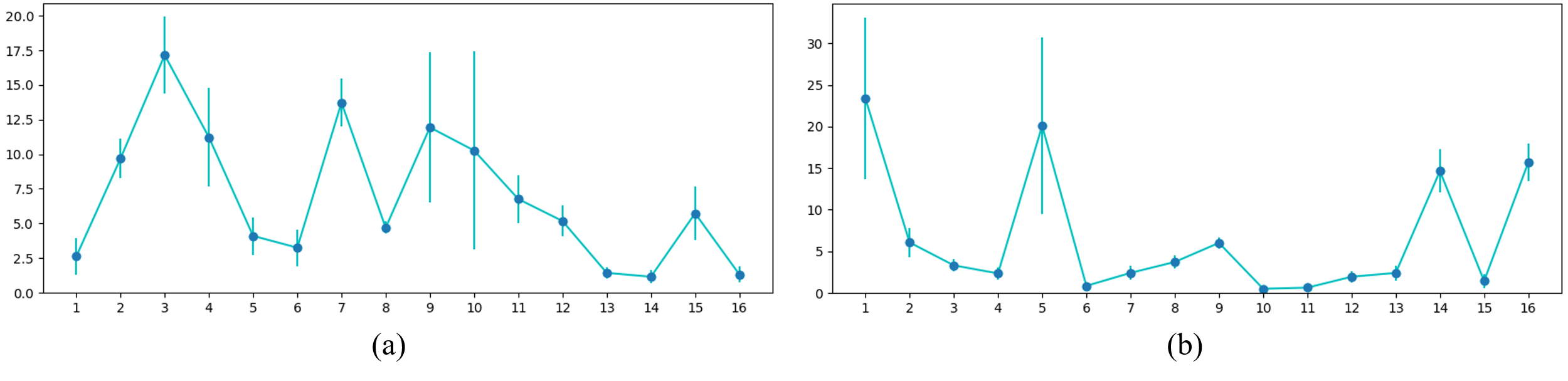}
  \vspace{-6mm}
  \captionsetup{font=footnotesize}
  \caption{{\textbf{Number of Objects per Frame.} (a) campus and (b) factory.}
  }
  \vspace{-3mm}
  \label{fig:num_obj}
\end{figure*}

\begin{figure*}[!h]
  \includegraphics[width=\linewidth]{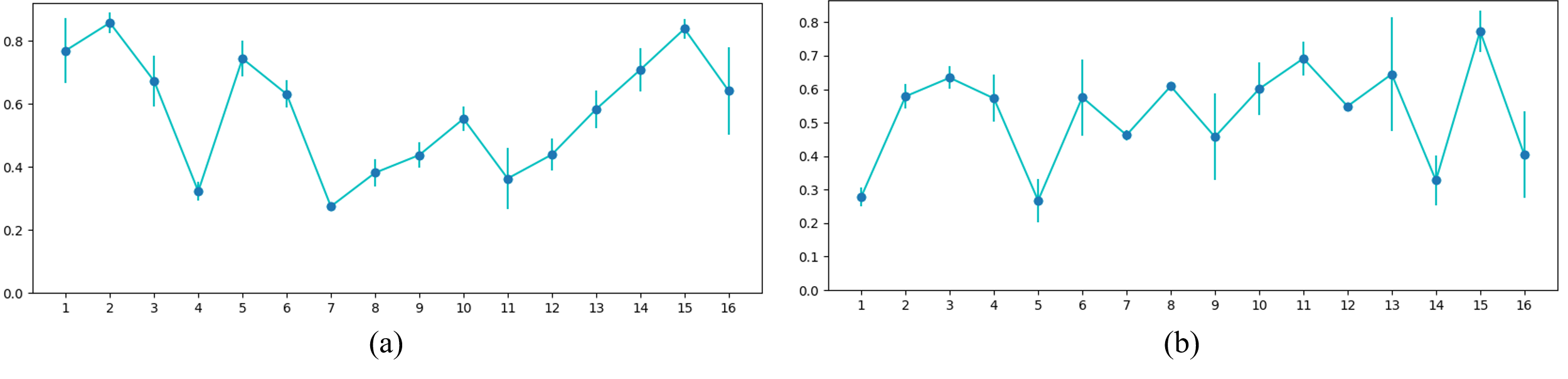}
  \vspace{-6mm}
  \captionsetup{font=footnotesize}
  \caption{{\textbf{Analysis on Association Performance.} For the analysis, we adopt AssA~\cite{luiten2021hota} with the localisation threshold $\alpha=50$.}
  }
  \vspace{-3mm}
  \label{fig:asso_anal}
\end{figure*}

\begin{figure*}[!h]
  \includegraphics[width=\linewidth]{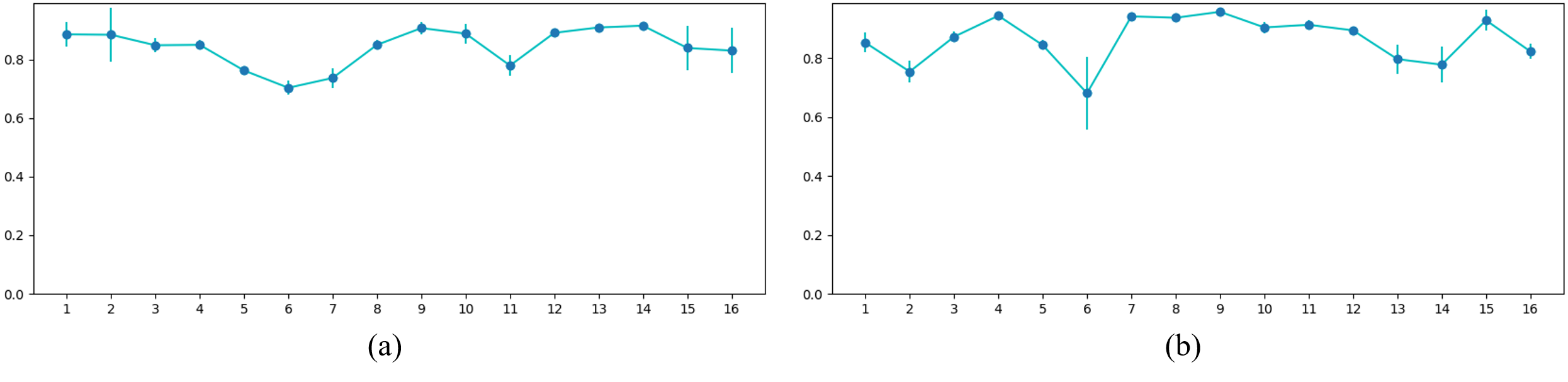}
  \vspace{-6mm}
  \captionsetup{font=footnotesize}
\caption{{\textbf{Analysis on Detection Performance.} For the analysis, we adopt DetA~\cite{luiten2021hota} with the localisation threshold $\alpha=50$.}
  }
  \vspace{-9mm}
  \label{fig:det_anal}
\end{figure*}

\paragraph{{Number of Objects per Frame}}
The average number of objects per frame at each site is shown in Figure \ref{fig:num_obj}, with error bars indicating standard deviation. Our dataset showcases a significant variation in the number of objects per camera, mirroring real-world environments characterized by challenges like occlusion and diverse movement patterns. A noteworthy observation is the correlation between object density and the complexity of association tasks (Figures \ref{fig:asso_anal}, \ref{fig:det_anal}), implying that higher object densities escalate the intricacies of tracking.

\begin{figure*}[!h]
  \includegraphics[width=\linewidth]{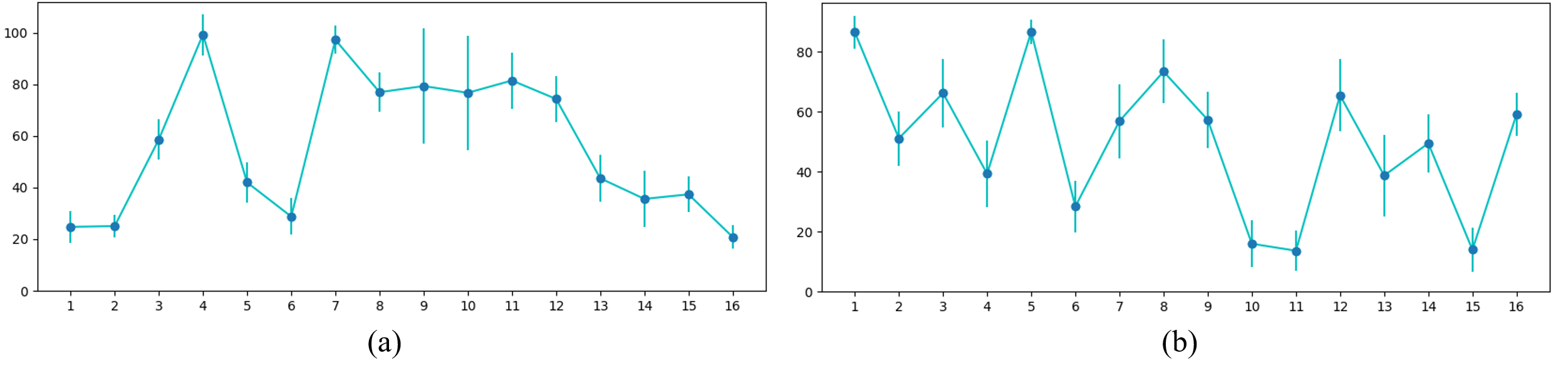}
  \vspace{-6mm}
  \captionsetup{font=footnotesize}
  \caption{{\textbf{Number of Tracks per Video.} (a) campus and (b) factory.}
  }
  \vspace{-3mm}
  \label{fig:num_track}
\end{figure*}

\paragraph{Number of Tracks per Video}
Figure \ref{fig:num_track} illustrates the average number of tracks per video. This measure is vital for understanding the range of tracking difficulties, demonstrating that complexity is influenced not only by the quantity of objects but also by the nature of multiple, distinct tracks. High track counts typically indicate more frequent interactions and overlapping paths, posing additional challenges and necessitating sophisticated algorithms for effective differentiation.

\begin{figure}[!t]
\begin{center}
\includegraphics[width=\linewidth]{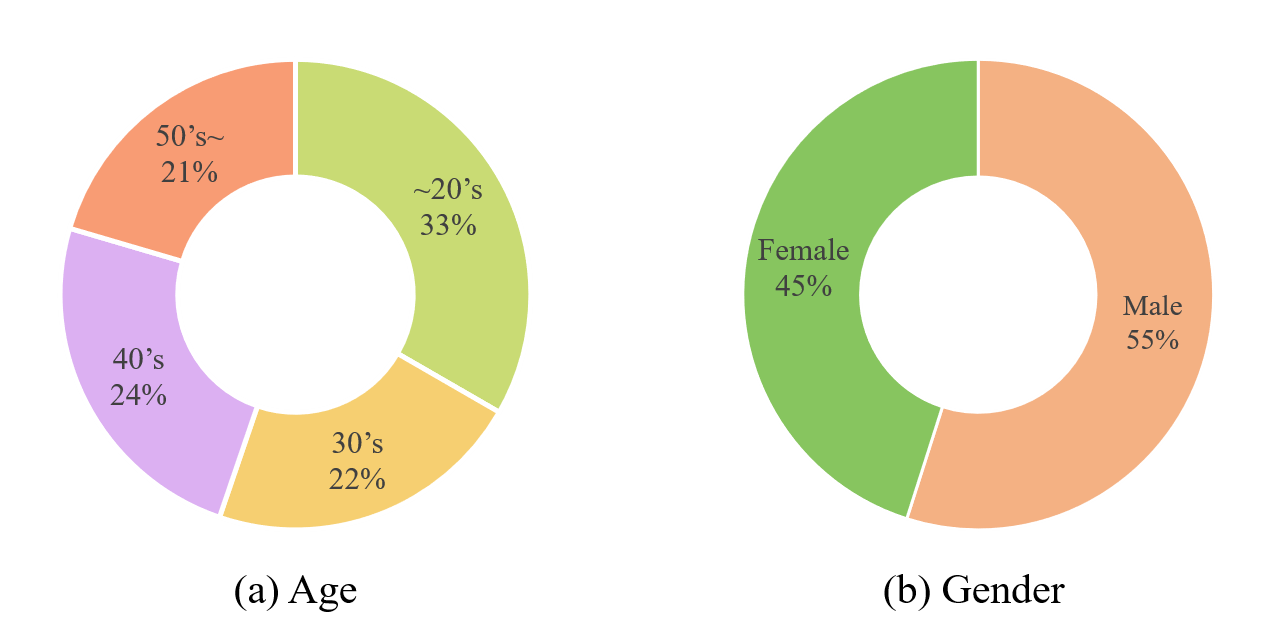}
\captionsetup{font=footnotesize}
\caption{\textbf{The age and gender distributions of actors.} 
}
\label{fig:actor}
\vspace{-6mm}
\end{center}
\end{figure}

\paragraph{The Age and Gender Distributions}
Figure \ref{fig:actor} shows the age and gender distribution of actors in our dataset. By encompassing a wide range of ages and genders, the dataset is relevant to diverse real-world settings. This variety does more than represent different demographics; it introduces added complexity to tracking tasks. Different movement and interaction patterns among various age groups and genders present additional challenges, particularly in dynamic or crowded settings.

\subsection{Full Illustrations}
\begin{figure*}[!h]
\begin{center}
\includegraphics[width=\textwidth,height=\textheight,keepaspectratio]{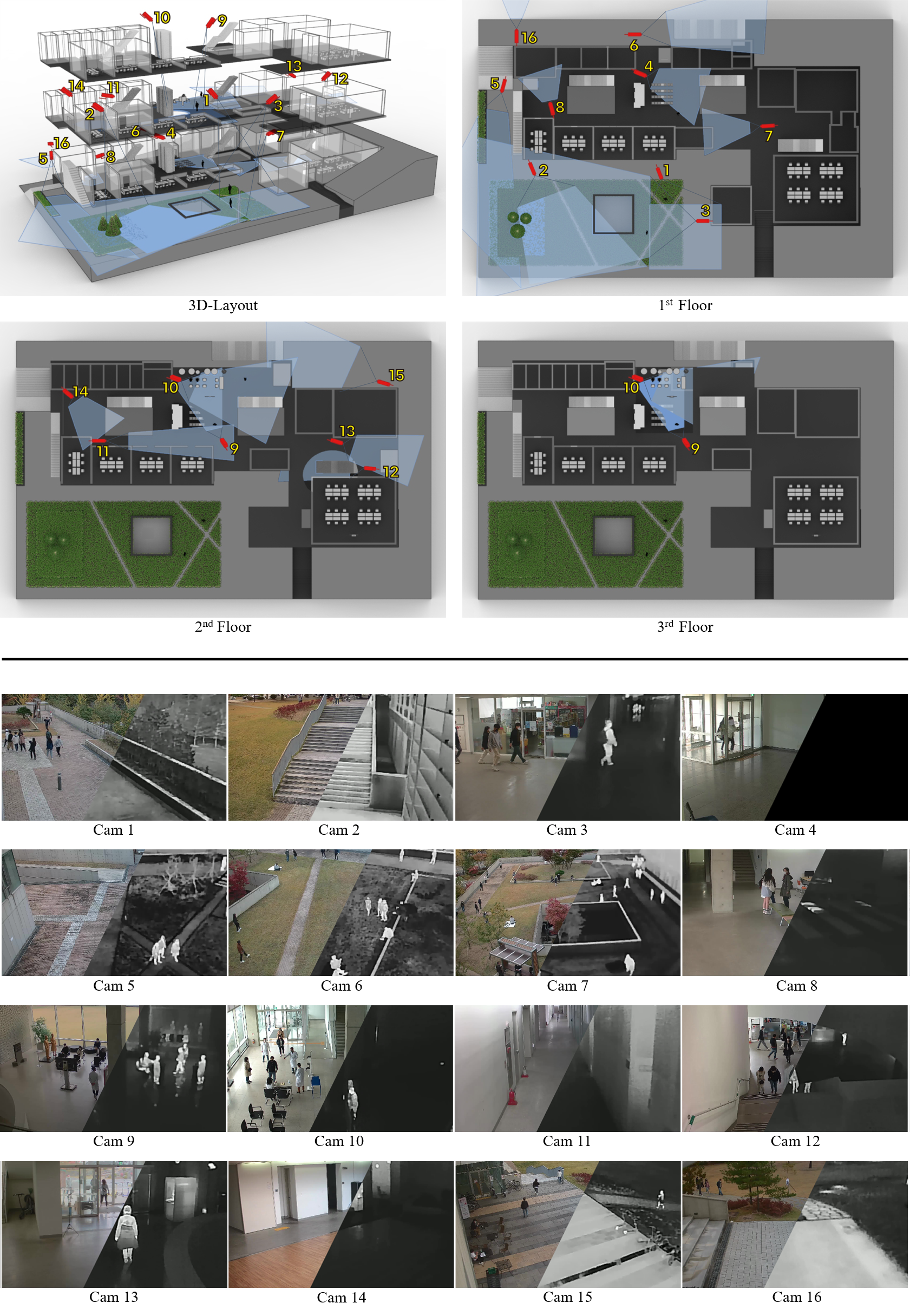}
\captionsetup{font=footnotesize}
\caption{(top) Camera layout on campus environment. (bottom) Examples of multi-spectral images on campus.} 
\label{fig:campus_supple}
\end{center}
\end{figure*}

\begin{figure*}[!h]
\begin{center}
\includegraphics[width=\textwidth,height=\textheight,keepaspectratio]{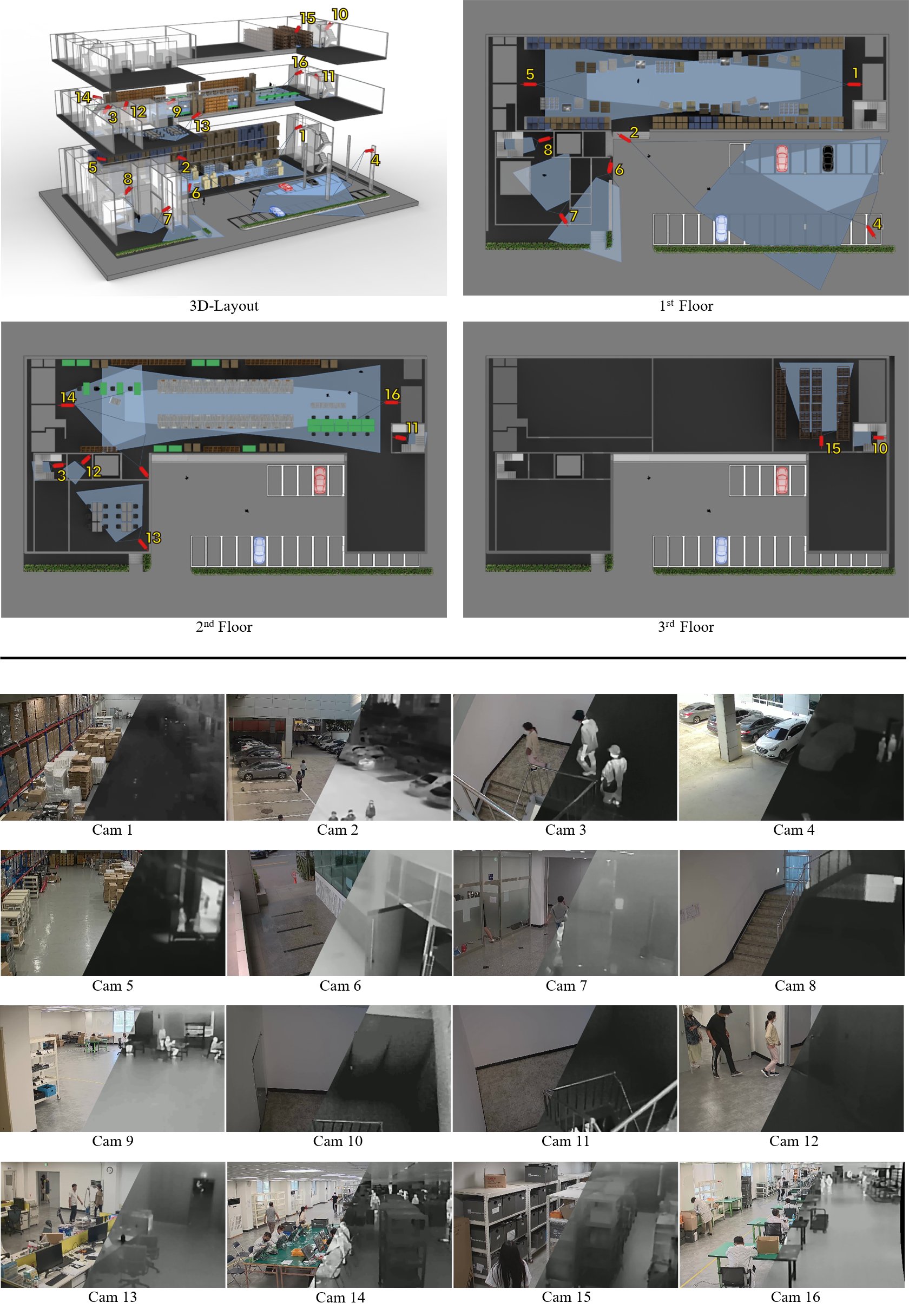}
\captionsetup{font=footnotesize}
\caption{(top) Camera layout on factory environment. (bottom) Examples of multi-spectral images on factory.
}
\label{fig:factory_supple}
\end{center}
\end{figure*}

We provide full illustrations of the cameras installed in the campus and factory environments in \figref{fig:campus_supple} and \figref{fig:factory_supple}, respectively. These cameras were installed in various locations, such as indoors, outdoors, and across different floors, replicating a dense real-world surveillance camera system.





\section{Dataset License}
The licenses of the datasets used in the experiments are denoted as follows:
\begin{itemize}\small\itemsep0.5em
  \item COCO2017~\cite{lin2014microsoft}: CC BY 4.0
  \item Market-1501~\cite{zheng2015scalable}: \url{https://zheng-lab.cecs.anu.edu.au/Project/project_reid.html} 
  \item MSMT17~\cite{wei2018person}: \url{https://www.pkuvmc.com/publications/msmt17.html}
  \item MOT17~\cite{milan2016mot16}: CC BY-NC-SA 3.0
  \item MOTSynth~\cite{fabbri2021motsynth}: CC BY-NC-SA 3.0
\end{itemize}

\section{Video Demo}
Our demo highlights the MTMMC dataset's distinct characteristics, including multi-modal data, extended and varied tracks, and complex scenarios. It features diverse situations recorded by RGB-Thermal cameras in campus and factory settings, illustrating the dataset's proficiency in tracking multiple targets across different camera perspectives, managing occlusions, and navigating challenging lighting conditions.


\section{Ethical Considerations}

In creating the MTMMC dataset, our foremost commitment is to the protection of personal privacy and ethical integrity in data usage. We meticulously selected school campuses and factory locations where we had comprehensive authorization, ensuring total control over the data collection process. This approach was governed by a stringent and transparent protocol. Each participant provided informed consent through release agreements, and we rigorously de-identified all non-actor data to further safeguard privacy. 
This meticulous process received the endorsement of the National Information Society Agency.
The Institutional Review Board (IRB) is presently conducting an in-depth review of the dataset, underscoring our dedication to ethical compliance. Furthermore, we are committed to adhering to all relevant privacy laws and regulations, ensuring the dataset aligns with the highest standards of data protection.



Our intention in releasing the MTMMC dataset is to foster advancements in multi-target multi-camera tracking research. We aim to facilitate the generation of open-source, transparent academic works, enhancing knowledge and understanding within the scholarly community. The dataset is strictly designated for non-commercial, public, and academic research, with specific use cases detailed in the accompanying agreement. We vigilantly prohibit any unauthorized use that diverges from these outlined purposes, particularly to avoid potential civil rights violations. In such instances, we will take decisive action in accordance with applicable laws.

To mitigate any inadvertent misuse by third-party groups, we have instituted robust measures:
\begin{itemize}[leftmargin=*]\itemsep.5em
\item Access to the MTMMC dataset is granted exclusively upon formal request. Prospective users must sign a comprehensive usage agreement, outlining their responsibilities and the ethical boundaries of data utilization.
\item We maintain a proactive stance in monitoring dataset usage. The author's affiliated institution(s) reserve the right to report any suspicious activities or individuals to law enforcement officials or regulatory bodies, particularly in cases of legal or regulatory transgressions.
\item In collaboration with law enforcement agencies, we will actively participate in investigations and legal actions against any illicit activities involving the dataset.
\item Regular audits and reviews of dataset usage will be conducted to ensure continuous adherence to ethical standards and privacy regulations.
\item We have established training programs for all dataset users, emphasizing the importance of ethical data handling and awareness of privacy implications.
\item A transparent feedback mechanism is in place, allowing users and observers to voice ethical concerns or report misuse. This facilitates a responsive and accountable approach to data governance.
\item Our ethical practices are not static; they are subject to ongoing evaluation and refinement, reflecting the evolving landscape of data privacy and ethical norms.
\item We engage with external ethical boards and committees, seeking their guidance and oversight in maintaining the ethical integrity of our dataset.
\end{itemize}

Our comprehensive approach to ethical data management reflects our unwavering commitment to upholding the highest standards of privacy and integrity in academic research. Through these measures, we strive to ensure the MTMMC dataset serves as a valuable and responsible resource for the research community.

\clearpage

\clearpage


\clearpage

\begin{figure*}[!h]
\begin{center}
\includegraphics[width=\textwidth,height=\textheight,keepaspectratio]{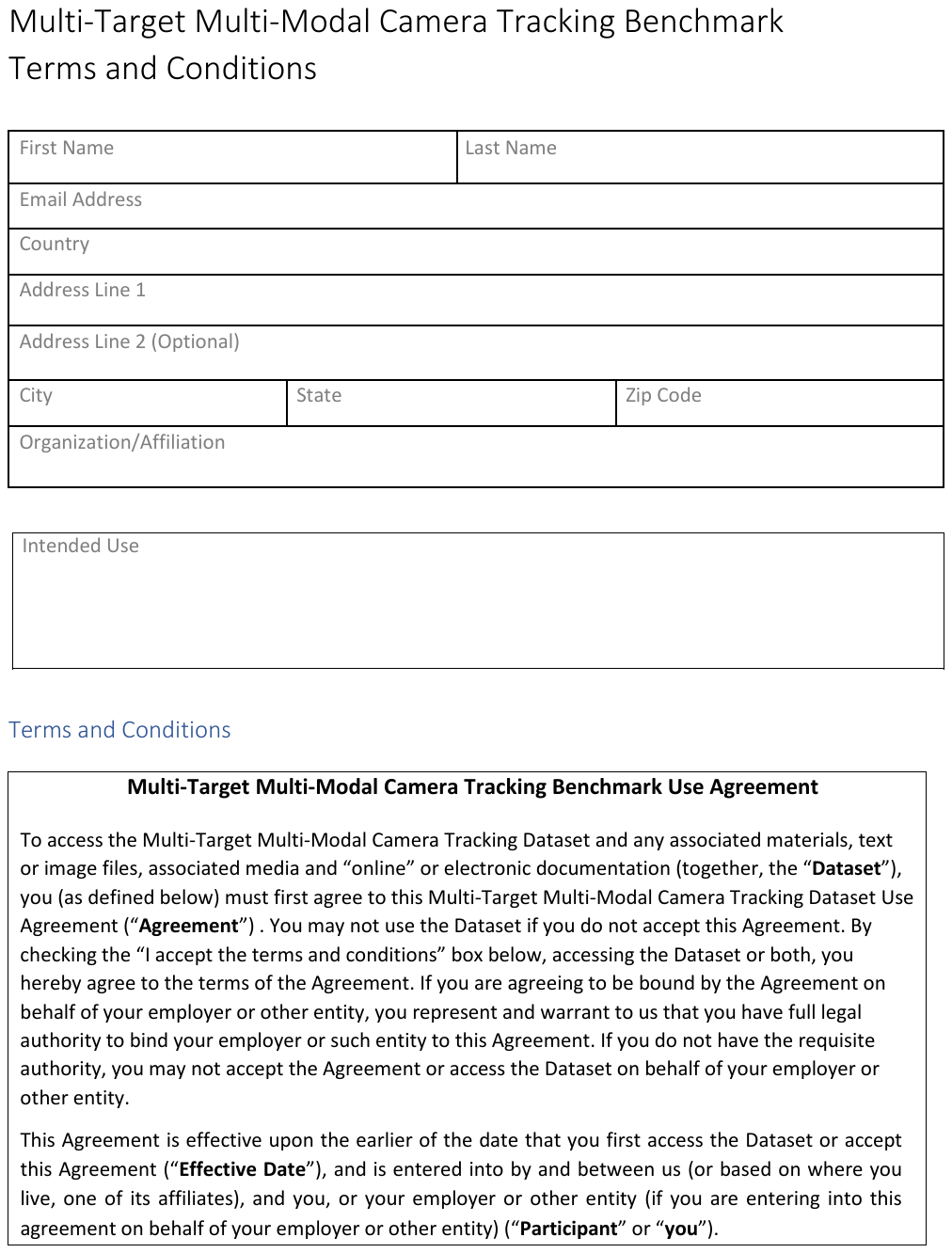}
\end{center}
\end{figure*}
\clearpage

\begin{figure*}[!h]
\begin{center}
\includegraphics[width=\textwidth,height=\textheight,keepaspectratio]{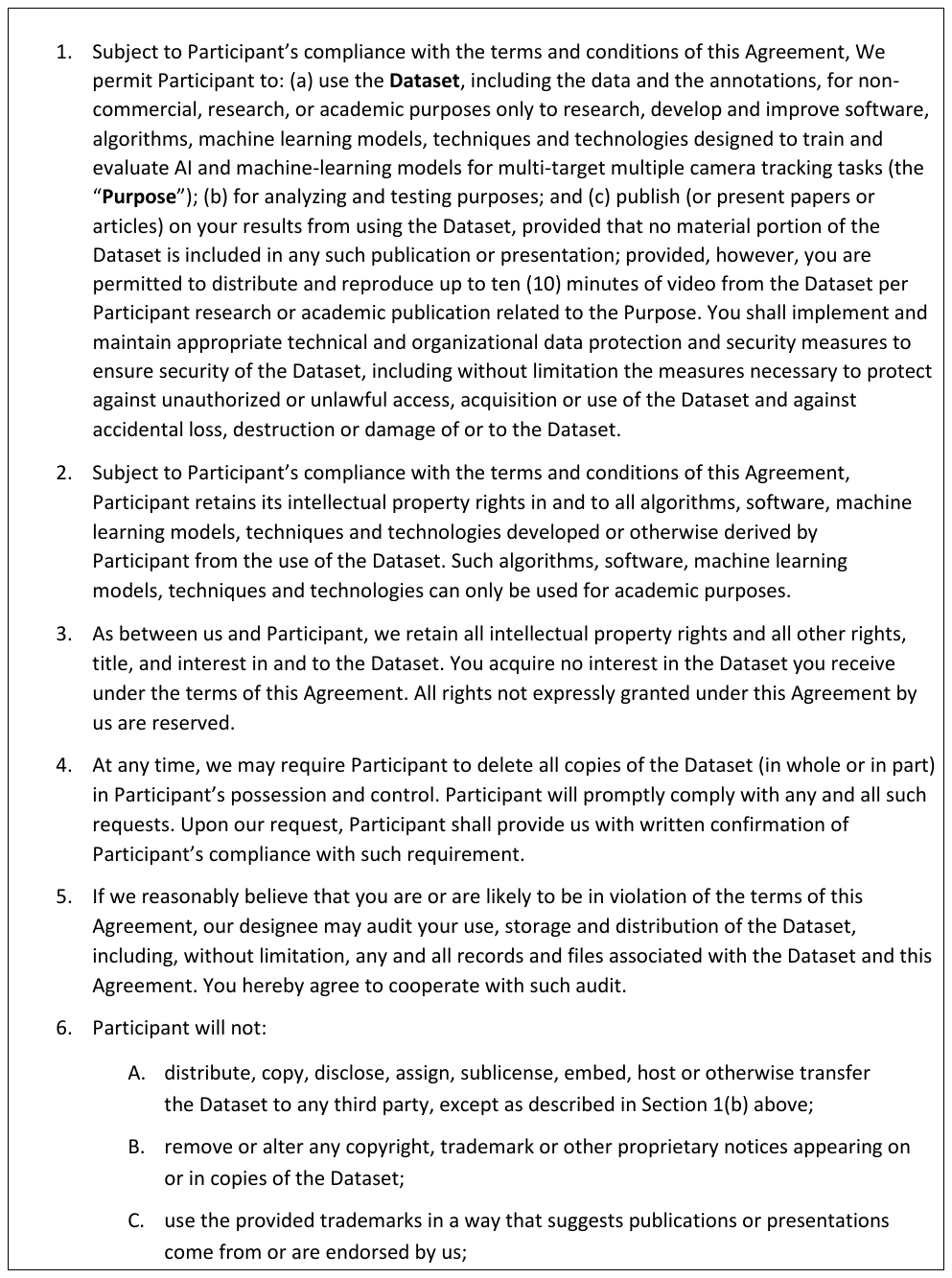}
\end{center}
\end{figure*}
\clearpage

\begin{figure*}[!h]
\begin{center}
\includegraphics[width=\textwidth,height=\textheight,keepaspectratio]{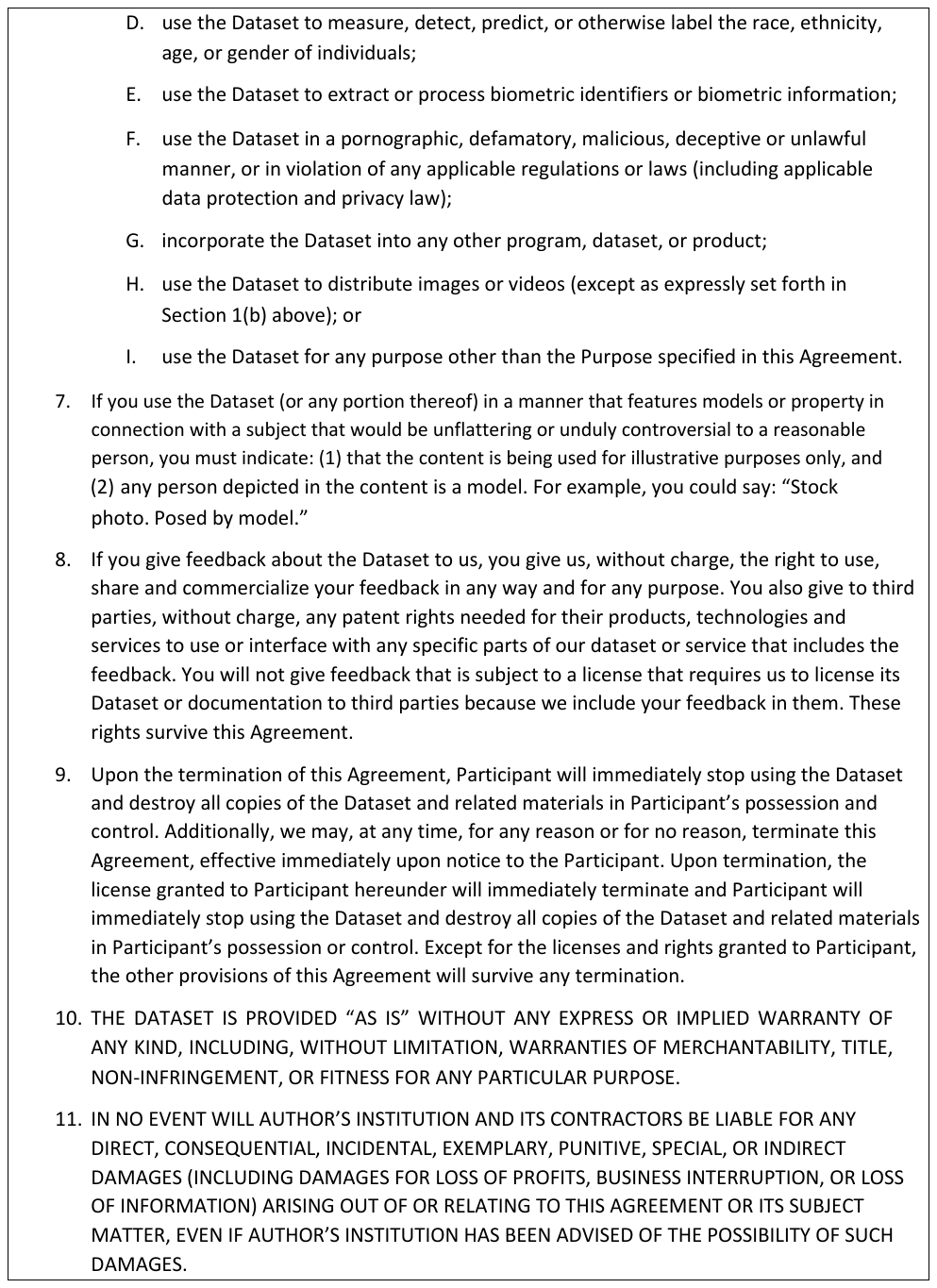}
\end{center}
\end{figure*}
\clearpage

\begin{figure*}[!h]
\begin{center}
\includegraphics[width=\textwidth,height=\textheight,keepaspectratio]{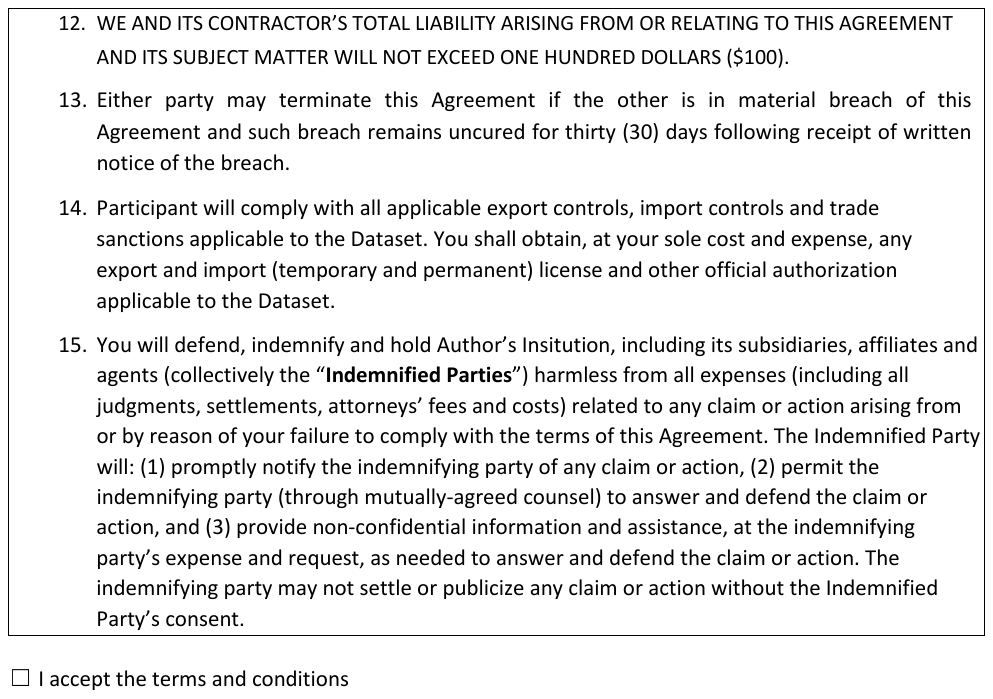}
\end{center}
\end{figure*}
\clearpage


{
    \small
    \bibliographystyle{ieeenat_fullname}
    \bibliography{main}

\begin{thebibliography}{94}
\providecommand{\natexlab}[1]{#1}
\providecommand{\url}[1]{\texttt{#1}}
\expandafter\ifx\csname urlstyle\endcsname\relax
  \providecommand{\doi}[1]{doi: #1}\else
  \providecommand{\doi}{doi: \begingroup \urlstyle{rm}\Url}\fi

\bibitem[Abavisani et~al.(2019)Abavisani, Joze, and Patel]{abavisani2019improving}
Mahdi Abavisani, Hamid Reza~Vaezi Joze, and Vishal~M Patel.
\newblock Improving the performance of unimodal dynamic hand-gesture recognition with multimodal training.
\newblock In \emph{Proceedings of the IEEE/CVF Conference on Computer Vision and Pattern Recognition}, pages 1165--1174, 2019.

\bibitem[Aharon et~al.(2022)Aharon, Orfaig, and Bobrovsky]{aharon2022bot}
Nir Aharon, Roy Orfaig, and Ben-Zion Bobrovsky.
\newblock Bot-sort: Robust associations multi-pedestrian tracking.
\newblock \emph{arXiv preprint arXiv:2206.14651}, 2022.

\bibitem[Batchuluun et~al.(2019)Batchuluun, Nguyen, Pham, Park, and Park]{batchuluun2019action}
Ganbayar Batchuluun, Dat~Tien Nguyen, Tuyen~Danh Pham, Chanhum Park, and Kang~Ryoung Park.
\newblock Action recognition from thermal videos.
\newblock \emph{IEEE Access}, 7:\penalty0 103893--103917, 2019.

\bibitem[Berclaz et~al.(2011)Berclaz, Fleuret, Turetken, and Fua]{berclaz2011multiple}
Jerome Berclaz, Francois Fleuret, Engin Turetken, and Pascal Fua.
\newblock Multiple object tracking using k-shortest paths optimization.
\newblock \emph{IEEE transactions on pattern analysis and machine intelligence}, 33\penalty0 (9):\penalty0 1806--1819, 2011.

\bibitem[Bonchi et~al.(2014)Bonchi, Garcia-Soriano, and Liberty]{bonchi2014correlation}
Francesco Bonchi, David Garcia-Soriano, and Edo Liberty.
\newblock Correlation clustering: from theory to practice.
\newblock In \emph{KDD}, page 1972, 2014.

\bibitem[Bredereck et~al.(2012)Bredereck, Jiang, K{\"o}rner, and Denzler]{bredereck2012data}
Michael Bredereck, Xiaoyan Jiang, Marco K{\"o}rner, and Joachim Denzler.
\newblock Data association for multi-object tracking-by-detection in multi-camera networks.
\newblock In \emph{2012 Sixth International Conference on Distributed Smart Cameras (ICDSC)}, pages 1--6. IEEE, 2012.

\bibitem[Cao et~al.(2023)Cao, Pang, Weng, Khirodkar, and Kitani]{cao2023observation}
Jinkun Cao, Jiangmiao Pang, Xinshuo Weng, Rawal Khirodkar, and Kris Kitani.
\newblock Observation-centric sort: Rethinking sort for robust multi-object tracking.
\newblock In \emph{Proceedings of the IEEE/CVF Conference on Computer Vision and Pattern Recognition}, pages 9686--9696, 2023.

\bibitem[Cao et~al.(2015)Cao, Chen, Chen, Zheng, and Huang]{cao2015equalised}
Lijun Cao, Weihua Chen, Xiaotang Chen, Shuai Zheng, and Kaiqi Huang.
\newblock An equalised global graphical model-based approach for multi-camera object tracking.
\newblock \emph{arXiv preprint arXiv:1502.03532}, 8, 2015.

\bibitem[Chavdarova et~al.(2017)Chavdarova, Baqu{\'e}, Bouquet, Maksai, Jose, Lettry, Fua, Van~Gool, and Fleuret]{chavdarova2017wildtrack}
Tatjana Chavdarova, Pierre Baqu{\'e}, St{\'e}phane Bouquet, Andrii Maksai, Cijo Jose, Louis Lettry, Pascal Fua, Luc Van~Gool, and Fran{\c{c}}ois Fleuret.
\newblock The wildtrack multi-camera person dataset.
\newblock \emph{arXiv preprint arXiv:1707.09299}, 2017.

\bibitem[Chen et~al.(2020)Chen, Ai, Chen, Zhuang, and Liu]{chen2020cross}
Long Chen, Haizhou Ai, Rui Chen, Zijie Zhuang, and Shuang Liu.
\newblock Cross-view tracking for multi-human 3d pose estimation at over 100 fps.
\newblock In \emph{Proceedings of the IEEE/CVF conference on computer vision and pattern recognition}, pages 3279--3288, 2020.

\bibitem[Contributors(2020)]{mmtrack2020}
MMTracking Contributors.
\newblock {MMTracking: OpenMMLab} video perception toolbox and benchmark.
\newblock \url{https://github.com/open-mmlab/mmtracking}, 2020.

\bibitem[Das et~al.(2014)Das, Chakraborty, and Roy-Chowdhury]{das2014consistent}
Abir Das, Anirban Chakraborty, and Amit~K Roy-Chowdhury.
\newblock Consistent re-identification in a camera network.
\newblock In \emph{European conference on computer vision}, pages 330--345. Springer, 2014.

\bibitem[Dave et~al.(2020)Dave, Khurana, Tokmakov, Schmid, and Ramanan]{dave2020tao}
Achal Dave, Tarasha Khurana, Pavel Tokmakov, Cordelia Schmid, and Deva Ramanan.
\newblock Tao: A large-scale benchmark for tracking any object.
\newblock In \emph{European conference on computer vision}, pages 436--454. Springer, 2020.

\bibitem[De~Vleeschouwer et~al.(2008)De~Vleeschouwer, Chen, Delannay, Parisot, Chaudy, Martrou, Cavallaro, et~al.]{de2008distributed}
Christophe De~Vleeschouwer, Fan Chen, Damien Delannay, Christophe Parisot, Christophe Chaudy, Eric Martrou, Andrea Cavallaro, et~al.
\newblock Distributed video acquisition and annotation for sport-event summarization.
\newblock \emph{NEM summit}, 8, 2008.

\bibitem[Deng et~al.(2019)Deng, Yang, Li, He, and Wang]{deng2019rfbnet}
Liuyuan Deng, Ming Yang, Tianyi Li, Yuesheng He, and Chunxiang Wang.
\newblock Rfbnet: deep multimodal networks with residual fusion blocks for rgb-d semantic segmentation.
\newblock \emph{arXiv preprint arXiv:1907.00135}, 2019.

\bibitem[Dollar et~al.(2009)Dollar, Wojek, Schiele, and Perona]{pedestrian}
Piotr Dollar, Christian Wojek, Bernt Schiele, and Pietro Perona.
\newblock Pedestrian detection: A benchmark.
\newblock In \emph{2009 IEEE Conference on Computer Vision and Pattern Recognition}, pages 304--311, 2009.

\bibitem[D'Orazio et~al.(2009)D'Orazio, Leo, Mosca, Spagnolo, and Mazzeo]{d2009semi}
Tiziana D'Orazio, Marco Leo, Nicola Mosca, Paolo Spagnolo, and Pier~Luigi Mazzeo.
\newblock A semi-automatic system for ground truth generation of soccer video sequences.
\newblock In \emph{2009 Sixth IEEE International Conference on Advanced Video and Signal Based Surveillance}, pages 559--564. IEEE, 2009.

\bibitem[Eigen and Fergus(2015)]{eigen2015predicting}
David Eigen and Rob Fergus.
\newblock Predicting depth, surface normals and semantic labels with a common multi-scale convolutional architecture.
\newblock In \emph{Proceedings of the IEEE international conference on computer vision}, pages 2650--2658, 2015.

\bibitem[Ess et~al.(2008)Ess, Leibe, Schindler, and Van~Gool]{eth_paper}
Andreas Ess, Bastian Leibe, Konrad Schindler, and Luc Van~Gool.
\newblock A mobile vision system for robust multi-person tracking.
\newblock In \emph{2008 IEEE Conference on Computer Vision and Pattern Recognition}, pages 1--8, 2008.

\bibitem[Fabbri et~al.(2021)Fabbri, Bras{\'o}, Maugeri, Cetintas, Gasparini, O{\v{s}}ep, Calderara, Leal-Taix{\'e}, and Cucchiara]{fabbri2021motsynth}
Matteo Fabbri, Guillem Bras{\'o}, Gianluca Maugeri, Orcun Cetintas, Riccardo Gasparini, Aljo{\v{s}}a O{\v{s}}ep, Simone Calderara, Laura Leal-Taix{\'e}, and Rita Cucchiara.
\newblock Motsynth: How can synthetic data help pedestrian detection and tracking?
\newblock In \emph{Proceedings of the IEEE/CVF International Conference on Computer Vision}, pages 10849--10859, 2021.

\bibitem[Feichtenhofer et~al.(2017)Feichtenhofer, Pinz, and Zisserman]{feichtenhofer2017detect}
Christoph Feichtenhofer, Axel Pinz, and Andrew Zisserman.
\newblock Detect to track and track to detect.
\newblock In \emph{Proceedings of the IEEE international conference on computer vision}, pages 3038--3046, 2017.

\bibitem[Felzenszwalb et~al.(2010)Felzenszwalb, Girshick, McAllester, and Ramanan]{felzenszwalb2010object}
Pedro~F Felzenszwalb, Ross~B Girshick, David McAllester, and Deva Ramanan.
\newblock Object detection with discriminatively trained part-based models.
\newblock \emph{IEEE transactions on pattern analysis and machine intelligence}, 32\penalty0 (9):\penalty0 1627--1645, 2010.

\bibitem[Ferryman and Shahrokni(2009)]{ferryman2009pets2009}
James Ferryman and Ali Shahrokni.
\newblock Pets2009: Dataset and challenge.
\newblock In \emph{2009 Twelfth IEEE international workshop on performance evaluation of tracking and surveillance}, pages 1--6. IEEE, 2009.

\bibitem[Fleuret et~al.(2007)Fleuret, Berclaz, Lengagne, and Fua]{fleuret2007multicamera}
Francois Fleuret, Jerome Berclaz, Richard Lengagne, and Pascal Fua.
\newblock Multicamera people tracking with a probabilistic occupancy map.
\newblock \emph{IEEE transactions on pattern analysis and machine intelligence}, 30\penalty0 (2):\penalty0 267--282, 2007.

\bibitem[Ge et~al.(2021)Ge, Liu, Wang, Li, and Sun]{ge2021yolox}
Zheng Ge, Songtao Liu, Feng Wang, Zeming Li, and Jian Sun.
\newblock Yolox: Exceeding yolo series in 2021.
\newblock \emph{arXiv preprint arXiv:2107.08430}, 2021.

\bibitem[Han et~al.(2021)Han, You, Wang, Zhang, Chu, Hu, Wang, and Liu]{han2021mmptrack}
Xiaotian Han, Quanzeng You, Chunyu Wang, Zhizheng Zhang, Peng Chu, Houdong Hu, Jiang Wang, and Zicheng Liu.
\newblock Mmptrack: Large-scale densely annotated multi-camera multiple people tracking benchmark.
\newblock \emph{arXiv preprint arXiv:2111.15157}, 2021.

\bibitem[He et~al.(2020{\natexlab{a}})He, Liao, Liu, Liu, Cheng, and Mei]{he2020fastreid}
Lingxiao He, Xingyu Liao, Wu Liu, Xinchen Liu, Peng Cheng, and Tao Mei.
\newblock Fastreid: A pytorch toolbox for general instance re-identification.
\newblock \emph{arXiv preprint arXiv:2006.02631}, 2020{\natexlab{a}}.

\bibitem[He et~al.(2020{\natexlab{b}})He, Wei, Hong, Shi, and Gong]{he2020multi}
Yuhang He, Xing Wei, Xiaopeng Hong, Weiwei Shi, and Yihong Gong.
\newblock Multi-target multi-camera tracking by tracklet-to-target assignment.
\newblock \emph{IEEE Transactions on Image Processing}, 29:\penalty0 5191--5205, 2020{\natexlab{b}}.

\bibitem[Hou et~al.(2021)Hou, Wang, Wang, and Zheng]{hou2021adaptive}
Yunzhong Hou, Zhongdao Wang, Shengjin Wang, and Liang Zheng.
\newblock Adaptive affinity for associations in multi-target multi-camera tracking.
\newblock \emph{IEEE Transactions on Image Processing}, 31:\penalty0 612--622, 2021.

\bibitem[Hu et~al.(2019)Hu, Cai, Wang, Lin, Sun, Krahenbuhl, Darrell, and Yu]{hu2019joint}
Hou-Ning Hu, Qi-Zhi Cai, Dequan Wang, Ji Lin, Min Sun, Philipp Krahenbuhl, Trevor Darrell, and Fisher Yu.
\newblock Joint monocular 3d vehicle detection and tracking.
\newblock In \emph{Proceedings of the IEEE/CVF International Conference on Computer Vision}, pages 5390--5399, 2019.

\bibitem[Hwang et~al.(2015)Hwang, Park, Kim, Choi, and So~Kweon]{hwang2015multispectral}
Soonmin Hwang, Jaesik Park, Namil Kim, Yukyung Choi, and In So~Kweon.
\newblock Multispectral pedestrian detection: Benchmark dataset and baseline.
\newblock In \emph{Proceedings of the IEEE conference on computer vision and pattern recognition}, pages 1037--1045, 2015.

\bibitem[Jiang et~al.(2018)Jiang, Bai, Xu, Xing, Zhou, and Wu]{jiang2018online}
Na Jiang, SiChen Bai, Yue Xu, Chang Xing, Zhong Zhou, and Wei Wu.
\newblock Online inter-camera trajectory association exploiting person re-identification and camera topology.
\newblock In \emph{Proceedings of the 26th ACM international conference on Multimedia}, pages 1457--1465, 2018.

\bibitem[Kart et~al.(2018)Kart, Kamarainen, and Matas]{kart2018make}
Ugur Kart, Joni-Kristian Kamarainen, and Jiri Matas.
\newblock How to make an rgbd tracker?
\newblock In \emph{proceedings of the european conference on computer vision (ECCV) Workshops}, pages 0--0, 2018.

\bibitem[Kart et~al.(2019)Kart, Lukezic, Kristan, Kamarainen, and Matas]{kart2019object}
Ugur Kart, Alan Lukezic, Matej Kristan, Joni-Kristian Kamarainen, and Jiri Matas.
\newblock Object tracking by reconstruction with view-specific discriminative correlation filters.
\newblock In \emph{Proceedings of the IEEE/CVF Conference on Computer Vision and Pattern Recognition}, pages 1339--1348, 2019.

\bibitem[Kohl et~al.(2020)Kohl, Specker, Schumann, and Beyerer]{kohl2020mta}
Philipp Kohl, Andreas Specker, Arne Schumann, and Jurgen Beyerer.
\newblock The mta dataset for multi-target multi-camera pedestrian tracking by weighted distance aggregation.
\newblock In \emph{Proceedings of the IEEE/CVF Conference on Computer Vision and Pattern Recognition Workshops}, pages 1042--1043, 2020.

\bibitem[Kristan et~al.(2020)Kristan, Leonardis, Matas, Felsberg, Pflugfelder, K{\"a}m{\"a}r{\"a}inen, Danelljan, Zajc, Luke{\v{z}}i{\v{c}}, Drbohlav, et~al.]{kristan2020eighth}
Matej Kristan, Ale{\v{s}} Leonardis, Ji{\v{r}}{\'\i} Matas, Michael Felsberg, Roman Pflugfelder, Joni-Kristian K{\"a}m{\"a}r{\"a}inen, Martin Danelljan, Luka~{\v{C}}ehovin Zajc, Alan Luke{\v{z}}i{\v{c}}, Ondrej Drbohlav, et~al.
\newblock The eighth visual object tracking vot2020 challenge results.
\newblock In \emph{Computer Vision--ECCV 2020 Workshops: Glasgow, UK, August 23--28, 2020, Proceedings, Part V 16}, pages 547--601. Springer, 2020.

\bibitem[Kuo et~al.(2010)Kuo, Huang, and Nevatia]{kuo2010inter}
Cheng-Hao Kuo, Chang Huang, and Ram Nevatia.
\newblock Inter-camera association of multi-target tracks by on-line learned appearance affinity models.
\newblock In \emph{European conference on computer vision}, pages 383--396. Springer, 2010.

\bibitem[Leal-Taix{\'e} et~al.(2016)Leal-Taix{\'e}, Canton-Ferrer, and Schindler]{leal2016learning}
Laura Leal-Taix{\'e}, Cristian Canton-Ferrer, and Konrad Schindler.
\newblock Learning by tracking: Siamese cnn for robust target association.
\newblock In \emph{Proceedings of the IEEE Conference on Computer Vision and Pattern Recognition Workshops}, pages 33--40, 2016.

\bibitem[Li et~al.(2019)Li, Song, Tong, and Tang]{li2019illumination}
Chengyang Li, Dan Song, Ruofeng Tong, and Min Tang.
\newblock Illumination-aware faster r-cnn for robust multispectral pedestrian detection.
\newblock \emph{Pattern Recognition}, 85:\penalty0 161--171, 2019.

\bibitem[Lin et~al.(2014)Lin, Maire, Belongie, Hays, Perona, Ramanan, Doll{\'a}r, and Zitnick]{lin2014microsoft}
Tsung-Yi Lin, Michael Maire, Serge Belongie, James Hays, Pietro Perona, Deva Ramanan, Piotr Doll{\'a}r, and C~Lawrence Zitnick.
\newblock Microsoft coco: Common objects in context.
\newblock In \emph{European conference on computer vision}, pages 740--755. Springer, 2014.

\bibitem[Lin et~al.(2017{\natexlab{a}})Lin, Doll{\'a}r, Girshick, He, Hariharan, and Belongie]{lin2017feature}
Tsung-Yi Lin, Piotr Doll{\'a}r, Ross Girshick, Kaiming He, Bharath Hariharan, and Serge Belongie.
\newblock Feature pyramid networks for object detection.
\newblock In \emph{Proceedings of the IEEE conference on computer vision and pattern recognition}, pages 2117--2125, 2017{\natexlab{a}}.

\bibitem[Lin et~al.(2017{\natexlab{b}})Lin, Goyal, Girshick, He, and Doll{\'a}r]{lin2017focal}
Tsung-Yi Lin, Priya Goyal, Ross Girshick, Kaiming He, and Piotr Doll{\'a}r.
\newblock Focal loss for dense object detection.
\newblock In \emph{Proceedings of the IEEE international conference on computer vision}, pages 2980--2988, 2017{\natexlab{b}}.

\bibitem[Liu et~al.(2018)Liu, Jing, Nie, Gao, Liu, and Jiang]{liu2018context}
Ye Liu, Xiao-Yuan Jing, Jianhui Nie, Hao Gao, Jun Liu, and Guo-Ping Jiang.
\newblock Context-aware three-dimensional mean-shift with occlusion handling for robust object tracking in rgb-d videos.
\newblock \emph{IEEE Transactions on Multimedia}, 21\penalty0 (3):\penalty0 664--677, 2018.

\bibitem[Luiten et~al.(2021)Luiten, Osep, Dendorfer, Torr, Geiger, Leal-Taix{\'e}, and Leibe]{luiten2021hota}
Jonathon Luiten, Aljosa Osep, Patrick Dendorfer, Philip Torr, Andreas Geiger, Laura Leal-Taix{\'e}, and Bastian Leibe.
\newblock Hota: A higher order metric for evaluating multi-object tracking.
\newblock \emph{International journal of computer vision}, 129\penalty0 (2):\penalty0 548--578, 2021.

\bibitem[Luo et~al.(2019)Luo, Gu, Liao, Lai, and Jiang]{luo2019bag}
Hao Luo, Youzhi Gu, Xingyu Liao, Shenqi Lai, and Wei Jiang.
\newblock Bag of tricks and a strong baseline for deep person re-identification.
\newblock In \emph{Proceedings of the IEEE/CVF conference on computer vision and pattern recognition workshops}, pages 0--0, 2019.

\bibitem[Luo et~al.(2018)Luo, Hsieh, Jiang, Niebles, and Fei-Fei]{luo2018graph}
Zelun Luo, Jun-Ting Hsieh, Lu Jiang, Juan~Carlos Niebles, and Li Fei-Fei.
\newblock Graph distillation for action detection with privileged modalities.
\newblock In \emph{Proceedings of the European Conference on Computer Vision (ECCV)}, pages 166--183, 2018.

\bibitem[Meinhardt et~al.(2021)Meinhardt, Kirillov, Leal-Taixe, and Feichtenhofer]{meinhardt2021trackformer}
Tim Meinhardt, Alexander Kirillov, Laura Leal-Taixe, and Christoph Feichtenhofer.
\newblock Trackformer: Multi-object tracking with transformers.
\newblock \emph{arXiv preprint arXiv:2101.02702}, 2021.

\bibitem[Milan et~al.(2016)Milan, Leal-Taix{\'e}, Reid, Roth, and Schindler]{milan2016mot16}
Anton Milan, Laura Leal-Taix{\'e}, Ian Reid, Stefan Roth, and Konrad Schindler.
\newblock Mot16: A benchmark for multi-object tracking.
\newblock \emph{arXiv preprint arXiv:1603.00831}, 2016.

\bibitem[Pang et~al.(2021)Pang, Qiu, Li, Chen, Li, Darrell, and Yu]{pang2021quasi}
Jiangmiao Pang, Linlu Qiu, Xia Li, Haofeng Chen, Qi Li, Trevor Darrell, and Fisher Yu.
\newblock Quasi-dense similarity learning for multiple object tracking.
\newblock In \emph{Proceedings of the IEEE/CVF conference on computer vision and pattern recognition}, pages 164--173, 2021.

\bibitem[Park et~al.(2018)Park, Woo, Lee, and Kweon]{park2018bam}
Jongchan Park, Sanghyun Woo, Joon-Young Lee, and In~So Kweon.
\newblock Bam: Bottleneck attention module.
\newblock \emph{arXiv preprint arXiv:1807.06514}, 2018.

\bibitem[Park et~al.(2022)Park, Woo, Oh, Kweon, and Lee]{park2022per}
Kwanyong Park, Sanghyun Woo, Seoung~Wug Oh, In~So Kweon, and Joon-Young Lee.
\newblock Per-clip video object segmentation.
\newblock In \emph{Proceedings of the IEEE/CVF Conference on Computer Vision and Pattern Recognition}, pages 1352--1361, 2022.

\bibitem[Quach et~al.(2021)Quach, Nguyen, Le, Truong, Duong, Tran, and Luu]{quach2021dyglip}
Kha~Gia Quach, Pha Nguyen, Huu Le, Thanh-Dat Truong, Chi~Nhan Duong, Minh-Triet Tran, and Khoa Luu.
\newblock Dyglip: A dynamic graph model with link prediction for accurate multi-camera multiple object tracking.
\newblock In \emph{Proceedings of the IEEE/CVF Conference on Computer Vision and Pattern Recognition}, pages 13784--13793, 2021.

\bibitem[Redmon and Farhadi(2017)]{redmon2017yolo9000}
Joseph Redmon and Ali Farhadi.
\newblock Yolo9000: better, faster, stronger.
\newblock In \emph{Proceedings of the IEEE conference on computer vision and pattern recognition}, pages 7263--7271, 2017.

\bibitem[Ren et~al.(2015)Ren, He, Girshick, and Sun]{ren2015faster}
Shaoqing Ren, Kaiming He, Ross Girshick, and Jian Sun.
\newblock Faster r-cnn: Towards real-time object detection with region proposal networks.
\newblock \emph{Advances in neural information processing systems}, 28, 2015.

\bibitem[Ristani and Tomasi(2018)]{ristani2018features}
Ergys Ristani and Carlo Tomasi.
\newblock Features for multi-target multi-camera tracking and re-identification.
\newblock In \emph{Proceedings of the IEEE conference on computer vision and pattern recognition}, pages 6036--6046, 2018.

\bibitem[Ristani et~al.(2016)Ristani, Solera, Zou, Cucchiara, and Tomasi]{ristani2016performance}
Ergys Ristani, Francesco Solera, Roger Zou, Rita Cucchiara, and Carlo Tomasi.
\newblock Performance measures and a data set for multi-target, multi-camera tracking.
\newblock In \emph{European conference on computer vision}, pages 17--35. Springer, 2016.

\bibitem[Shao et~al.(2018)Shao, Zhao, Li, Xiao, Yu, Zhang, and Sun]{shao2018crowdhuman}
Shuai Shao, Zijian Zhao, Boxun Li, Tete Xiao, Gang Yu, Xiangyu Zhang, and Jian Sun.
\newblock Crowdhuman: A benchmark for detecting human in a crowd.
\newblock \emph{arXiv preprint arXiv:1805.00123}, 2018.

\bibitem[Shin et~al.(2023)Shin, Park, Lee, Lee, and Kweon]{shin2023self}
Ukcheol Shin, Kwanyong Park, Byeong-Uk Lee, Kyunghyun Lee, and In~So Kweon.
\newblock Self-supervised monocular depth estimation from thermal images via adversarial multi-spectral adaptation.
\newblock In \emph{Proceedings of the IEEE/CVF Winter Conference on Applications of Computer Vision}, pages 5798--5807, 2023.

\bibitem[Shiva~Kumar et~al.(2017)Shiva~Kumar, Ramakrishnan, and Rathna]{shiva2017distributed}
KA Shiva~Kumar, KR Ramakrishnan, and GN Rathna.
\newblock Distributed person of interest tracking in camera networks.
\newblock In \emph{Proceedings of the 11th International Conference on Distributed Smart Cameras}, pages 131--137, 2017.

\bibitem[Sun et~al.(2020{\natexlab{a}})Sun, Cao, Jiang, Zhang, Xie, Yuan, Wang, and Luo]{sun2020transtrack}
Peize Sun, Jinkun Cao, Yi Jiang, Rufeng Zhang, Enze Xie, Zehuan Yuan, Changhu Wang, and Ping Luo.
\newblock Transtrack: Multiple object tracking with transformer.
\newblock \emph{arXiv preprint arXiv:2012.15460}, 2020{\natexlab{a}}.

\bibitem[Sun et~al.(2020{\natexlab{b}})Sun, Kretzschmar, Dotiwalla, Chouard, Patnaik, Tsui, Guo, Zhou, Chai, Caine, et~al.]{sun2020scalability}
Pei Sun, Henrik Kretzschmar, Xerxes Dotiwalla, Aurelien Chouard, Vijaysai Patnaik, Paul Tsui, James Guo, Yin Zhou, Yuning Chai, Benjamin Caine, et~al.
\newblock Scalability in perception for autonomous driving: Waymo open dataset.
\newblock In \emph{Proceedings of the IEEE/CVF conference on computer vision and pattern recognition}, pages 2446--2454, 2020{\natexlab{b}}.

\bibitem[Sun et~al.(2019{\natexlab{a}})Sun, Akhtar, Song, Mian, and Shah]{sun2019deep}
ShiJie Sun, Naveed Akhtar, HuanSheng Song, Ajmal Mian, and Mubarak Shah.
\newblock Deep affinity network for multiple object tracking.
\newblock \emph{IEEE transactions on pattern analysis and machine intelligence}, 43\penalty0 (1):\penalty0 104--119, 2019{\natexlab{a}}.

\bibitem[Sun et~al.(2019{\natexlab{b}})Sun, Zuo, and Liu]{sun2019rtfnet}
Yuxiang Sun, Weixun Zuo, and Ming Liu.
\newblock Rtfnet: Rgb-thermal fusion network for semantic segmentation of urban scenes.
\newblock \emph{IEEE Robotics and Automation Letters}, 4\penalty0 (3):\penalty0 2576--2583, 2019{\natexlab{b}}.

\bibitem[Van~Zandycke et~al.(2022)Van~Zandycke, Somers, Istasse, Don, and Zambrano]{van2022deepsportradar}
Gabriel Van~Zandycke, Vladimir Somers, Maxime Istasse, Carlo~Del Don, and Davide Zambrano.
\newblock Deepsportradar-v1: Computer vision dataset for sports understanding with high quality annotations.
\newblock In \emph{Proceedings of the 5th International ACM Workshop on Multimedia Content Analysis in Sports}, pages 1--8, 2022.

\bibitem[Wang et~al.(2021)Wang, Feiszli, Wang, and Tran]{wang2021unidentified}
Weiyao Wang, Matt Feiszli, Heng Wang, and Du Tran.
\newblock Unidentified video objects: A benchmark for dense, open-world segmentation.
\newblock In \emph{Proceedings of the IEEE/CVF International Conference on Computer Vision}, pages 10776--10785, 2021.

\bibitem[Wang et~al.(2020{\natexlab{a}})Wang, Zheng, Liu, Li, and Wang]{wang2020towards}
Zhongdao Wang, Liang Zheng, Yixuan Liu, Yali Li, and Shengjin Wang.
\newblock Towards real-time multi-object tracking.
\newblock In \emph{European Conference on Computer Vision}, pages 107--122. Springer, 2020{\natexlab{a}}.

\bibitem[Wang et~al.(2020{\natexlab{b}})Wang, Zheng, Liu, and Wang]{wang2019towards}
Zhongdao Wang, Liang Zheng, Yixuan Liu, and Shengjin Wang.
\newblock Towards real-time multi-object tracking.
\newblock \emph{The European Conference on Computer Vision (ECCV)}, 2020{\natexlab{b}}.

\bibitem[Weber et~al.(2021)Weber, Xie, Collins, Zhu, Voigtlaender, Adam, Green, Geiger, Leibe, Cremers, et~al.]{weber2021step}
Mark Weber, Jun Xie, Maxwell Collins, Yukun Zhu, Paul Voigtlaender, Hartwig Adam, Bradley Green, Andreas Geiger, Bastian Leibe, Daniel Cremers, et~al.
\newblock Step: Segmenting and tracking every pixel.
\newblock \emph{arXiv preprint arXiv:2102.11859}, 2021.

\bibitem[Wei et~al.(2018)Wei, Zhang, Gao, and Tian]{wei2018person}
Longhui Wei, Shiliang Zhang, Wen Gao, and Qi Tian.
\newblock Person transfer gan to bridge domain gap for person re-identification.
\newblock In \emph{Proceedings of the IEEE conference on computer vision and pattern recognition}, pages 79--88, 2018.

\bibitem[Wojke et~al.(2017)Wojke, Bewley, and Paulus]{wojke2017simple}
Nicolai Wojke, Alex Bewley, and Dietrich Paulus.
\newblock Simple online and realtime tracking with a deep association metric.
\newblock In \emph{2017 IEEE international conference on image processing (ICIP)}, pages 3645--3649. IEEE, 2017.

\bibitem[Woo et~al.(2022{\natexlab{a}})Woo, Park, Oh, Kweon, and Lee]{woo2022bridging}
Sanghyun Woo, Kwanyong Park, Seoung~Wug Oh, In~So Kweon, and Joon-Young Lee.
\newblock Bridging images and videos: A simple learning framework for large vocabulary video object detection.
\newblock In \emph{European Conference on Computer Vision}, pages 238--258. Springer, 2022{\natexlab{a}}.

\bibitem[Woo et~al.(2022{\natexlab{b}})Woo, Park, Oh, Kweon, and Lee]{woo2022tracking}
Sanghyun Woo, Kwanyong Park, Seoung~Wug Oh, In~So Kweon, and Joon-Young Lee.
\newblock Tracking by associating clips.
\newblock In \emph{European Conference on Computer Vision}, pages 129--145. Springer, 2022{\natexlab{b}}.

\bibitem[Wu et~al.(2021)Wu, Cao, Song, Wang, Yang, and Yuan]{wu2021track}
Jialian Wu, Jiale Cao, Liangchen Song, Yu Wang, Ming Yang, and Junsong Yuan.
\newblock Track to detect and segment: An online multi-object tracker.
\newblock In \emph{Proceedings of the IEEE/CVF conference on computer vision and pattern recognition}, pages 12352--12361, 2021.

\bibitem[Xiao et~al.(2017)Xiao, Li, Wang, Lin, and Wang]{xiao2017joint}
Tong Xiao, Shuang Li, Bochao Wang, Liang Lin, and Xiaogang Wang.
\newblock Joint detection and identification feature learning for person search.
\newblock In \emph{Proceedings of the IEEE conference on computer vision and pattern recognition}, pages 3415--3424, 2017.

\bibitem[Xu et~al.(2017)Xu, Ouyang, Ricci, Wang, and Sebe]{xu2017learning}
Dan Xu, Wanli Ouyang, Elisa Ricci, Xiaogang Wang, and Nicu Sebe.
\newblock Learning cross-modal deep representations for robust pedestrian detection.
\newblock In \emph{Proceedings of the IEEE conference on computer vision and pattern recognition}, pages 5363--5371, 2017.

\bibitem[Yan et~al.(2021)Yan, Yang, K{\"a}pyl{\"a}, Zheng, Leonardis, and K{\"a}m{\"a}r{\"a}inen]{yan2021depthtrack}
Song Yan, Jinyu Yang, Jani K{\"a}pyl{\"a}, Feng Zheng, Ale{\v{s}} Leonardis, and Joni-Kristian K{\"a}m{\"a}r{\"a}inen.
\newblock Depthtrack: Unveiling the power of rgbd tracking.
\newblock In \emph{Proceedings of the IEEE/CVF International Conference on Computer Vision}, pages 10725--10733, 2021.

\bibitem[Ye et~al.(2021)Ye, Shen, Lin, Xiang, Shao, and Hoi]{ye2021deep}
Mang Ye, Jianbing Shen, Gaojie Lin, Tao Xiang, Ling Shao, and Steven~CH Hoi.
\newblock Deep learning for person re-identification: A survey and outlook.
\newblock \emph{IEEE transactions on pattern analysis and machine intelligence}, 44\penalty0 (6):\penalty0 2872--2893, 2021.

\bibitem[You and Jiang(2020)]{you2020real}
Quanzeng You and Hao Jiang.
\newblock Real-time 3d deep multi-camera tracking.
\newblock \emph{arXiv preprint arXiv:2003.11753}, 2020.

\bibitem[Yu et~al.(2020)Yu, Chen, Wang, Xian, Chen, Liu, Madhavan, and Darrell]{yu2020bdd100k}
Fisher Yu, Haofeng Chen, Xin Wang, Wenqi Xian, Yingying Chen, Fangchen Liu, Vashisht Madhavan, and Trevor Darrell.
\newblock Bdd100k: A diverse driving dataset for heterogeneous multitask learning.
\newblock In \emph{Proceedings of the IEEE/CVF conference on computer vision and pattern recognition}, pages 2636--2645, 2020.

\bibitem[Zeng et~al.(2021)Zeng, Dong, Wang, Zhang, and Wei]{zeng2021motr}
Fangao Zeng, Bin Dong, Tiancai Wang, Xiangyu Zhang, and Yichen Wei.
\newblock Motr: End-to-end multiple-object tracking with transformer.
\newblock \emph{arXiv preprint arXiv:2105.03247}, 2021.

\bibitem[Zhang et~al.(2019{\natexlab{a}})Zhang, Liu, Zhang, Yang, Qiao, Huang, and Hussain]{zhang2019cross}
Lu Zhang, Zhiyong Liu, Shifeng Zhang, Xu Yang, Hong Qiao, Kaizhu Huang, and Amir Hussain.
\newblock Cross-modality interactive attention network for multispectral pedestrian detection.
\newblock \emph{Information Fusion}, 50:\penalty0 20--29, 2019{\natexlab{a}}.

\bibitem[Zhang et~al.(2019{\natexlab{b}})Zhang, Zhu, Chen, Yang, Lei, and Liu]{zhang2019weakly}
Lu Zhang, Xiangyu Zhu, Xiangyu Chen, Xu Yang, Zhen Lei, and Zhiyong Liu.
\newblock Weakly aligned cross-modal learning for multispectral pedestrian detection.
\newblock In \emph{Proceedings of the IEEE/CVF International Conference on Computer Vision}, pages 5127--5137, 2019{\natexlab{b}}.

\bibitem[Zhang et~al.(2015)Zhang, Staudt, Faltemier, and Roy-Chowdhury]{zhang2015camera}
Shu Zhang, Elliot Staudt, Tim Faltemier, and Amit~K Roy-Chowdhury.
\newblock A camera network tracking (camnet) dataset and performance baseline.
\newblock In \emph{2015 IEEE Winter Conference on Applications of Computer Vision}, pages 365--372. IEEE, 2015.

\bibitem[Zhang et~al.(2017{\natexlab{a}})Zhang, Benenson, and Schiele]{zhang2017citypersons}
Shanshan Zhang, Rodrigo Benenson, and Bernt Schiele.
\newblock Citypersons: A diverse dataset for pedestrian detection.
\newblock In \emph{Proceedings of the IEEE conference on computer vision and pattern recognition}, pages 3213--3221, 2017{\natexlab{a}}.

\bibitem[Zhang et~al.(2021)Zhang, Sun, Jiang, Yu, Yuan, Luo, Liu, and Wang]{zhang2021bytetrack}
Yifu Zhang, Peize Sun, Yi Jiang, Dongdong Yu, Zehuan Yuan, Ping Luo, Wenyu Liu, and Xinggang Wang.
\newblock Bytetrack: Multi-object tracking by associating every detection box.
\newblock \emph{arXiv preprint arXiv:2110.06864}, 2021.

\bibitem[Zhang et~al.(2017{\natexlab{b}})Zhang, Wu, Zhang, and Zhang]{zhang2017multi}
Zhimeng Zhang, Jianan Wu, Xuan Zhang, and Chi Zhang.
\newblock Multi-target, multi-camera tracking by hierarchical clustering: Recent progress on dukemtmc project.
\newblock \emph{arXiv preprint arXiv:1712.09531}, 2017{\natexlab{b}}.

\bibitem[Zhao et~al.(2020)Zhao, Peng, Chen, Kapadia, and Metaxas]{zhao2020knowledge}
Long Zhao, Xi Peng, Yuxiao Chen, Mubbasir Kapadia, and Dimitris~N Metaxas.
\newblock Knowledge as priors: Cross-modal knowledge generalization for datasets without superior knowledge.
\newblock In \emph{Proceedings of the IEEE/CVF Conference on Computer Vision and Pattern Recognition}, pages 6528--6537, 2020.

\bibitem[Zheng et~al.(2015)Zheng, Shen, Tian, Wang, Wang, and Tian]{zheng2015scalable}
Liang Zheng, Liyue Shen, Lu Tian, Shengjin Wang, Jingdong Wang, and Qi Tian.
\newblock Scalable person re-identification: A benchmark.
\newblock In \emph{Proceedings of the IEEE international conference on computer vision}, pages 1116--1124, 2015.

\bibitem[Zheng et~al.(2017{\natexlab{a}})Zheng, Zhang, Sun, Chandraker, Yang, and Tian]{zheng2017person}
Liang Zheng, Hengheng Zhang, Shaoyan Sun, Manmohan Chandraker, Yi Yang, and Qi Tian.
\newblock Person re-identification in the wild.
\newblock In \emph{Proceedings of the IEEE Conference on Computer Vision and Pattern Recognition}, pages 1367--1376, 2017{\natexlab{a}}.

\bibitem[Zheng et~al.(2017{\natexlab{b}})Zheng, Zheng, and Yang]{zheng2017unlabeled}
Zhedong Zheng, Liang Zheng, and Yi Yang.
\newblock Unlabeled samples generated by gan improve the person re-identification baseline in vitro.
\newblock In \emph{Proceedings of the IEEE international conference on computer vision}, pages 3754--3762, 2017{\natexlab{b}}.

\bibitem[Zhong et~al.(2017)Zhong, Zheng, Cao, and Li]{zhong2017re}
Zhun Zhong, Liang Zheng, Donglin Cao, and Shaozi Li.
\newblock Re-ranking person re-identification with k-reciprocal encoding.
\newblock In \emph{Proceedings of the IEEE conference on computer vision and pattern recognition}, pages 1318--1327, 2017.

\bibitem[Zhou et~al.(2021{\natexlab{a}})Zhou, Lin, Lei, Yu, and Hwang]{zhou2021mffenet}
Wujie Zhou, Xinyang Lin, Jingsheng Lei, Lu Yu, and Jenq-Neng Hwang.
\newblock Mffenet: Multiscale feature fusion and enhancement network for rgb--thermal urban road scene parsing.
\newblock \emph{IEEE Transactions on Multimedia}, 24:\penalty0 2526--2538, 2021{\natexlab{a}}.

\bibitem[Zhou et~al.(2021{\natexlab{b}})Zhou, Liu, Lei, Yu, and Hwang]{zhou2021gmnet}
Wujie Zhou, Jinfu Liu, Jingsheng Lei, Lu Yu, and Jenq-Neng Hwang.
\newblock Gmnet: graded-feature multilabel-learning network for rgb-thermal urban scene semantic segmentation.
\newblock \emph{IEEE Transactions on Image Processing}, 30:\penalty0 7790--7802, 2021{\natexlab{b}}.

\bibitem[Zhou et~al.(2020)Zhou, Koltun, and Kr{\"a}henb{\"u}hl]{zhou2020tracking}
Xingyi Zhou, Vladlen Koltun, and Philipp Kr{\"a}henb{\"u}hl.
\newblock Tracking objects as points.
\newblock In \emph{European Conference on Computer Vision}, pages 474--490. Springer, 2020.

\end{thebibliography}
}


\end{document}